%% file: manuscript-main.tex
\documentclass{article}

\usepackage{manuscript-main}

\usepackage[utf8]{inputenc} % allow utf-8 input
\usepackage[T1]{fontenc}    % use 8-bit T1 fonts

\usepackage{charter}
\usepackage[scaled=0.92]{helvet}

\usepackage{hyperref}       % hyperlinks
\usepackage{url}            % simple URL typesetting
\usepackage{booktabs}       % professional-quality tables
\usepackage{array}
\usepackage{amsmath}        % math environments and \boldsymbol
\usepackage{amsfonts}       % blackboard math symbols
\usepackage{nicefrac}       % compact symbols for 1/2, etc.
\usepackage{microtype}      % microtypography
\usepackage{lipsum}
\usepackage{fancyhdr}       % header
\usepackage{graphicx}       % graphics
\usepackage{multirow}
\usepackage{xcolor}
\graphicspath{{media/}}     % organize your images and other figures under media/ folder
\usepackage{tabularx}

\usepackage[font=small,labelfont=bf]{caption}

\definecolor{citelinkblue}{HTML}{0056D6}
\hypersetup{citecolor=citelinkblue, linkcolor=citelinkblue}

\newcommand{\figref}[1]{\hyperref[#1]{Fig.~\ref*{#1}}}
\newcommand{\figsref}[1]{\hyperref[#1]{Figs.~\ref*{#1}}}
\newcommand{\tabref}[1]{\hyperref[#1]{Table~\ref*{#1}}}
\newcommand{\eqnref}[1]{\hyperref[#1]{Eq.~\ref*{#1}}}
\newcommand{\secref}[1]{\hyperref[#1]{Section~\ref*{#1}}}
\newcommand{\appref}[1]{\hyperref[#1]{Appendix~\ref*{#1}}}

\usepackage{setspace}

\usepackage{etoolbox}
\AtBeginEnvironment{figure}{\singlespacing}
\AtBeginEnvironment{table}{\singlespacing\small}
\AtBeginEnvironment{tabular}{\small}
\AtBeginEnvironment{tabularx}{\small}

\makeatletter
\def\hyper@natlinkbreak#1#2{#1}
\patchcmd{\NAT@citex}
  {\@citea\NAT@hyper@{%
     \NAT@nmfmt{\NAT@nm}%
     \hyper@natlinkbreak{\NAT@spacechar\NAT@@open\if*#1*\else#1\NAT@spacechar\fi}%
       {\@citeb\@extra@b@citeb}%
     \NAT@date}}
  {\@citea\NAT@hyper@{%
     \NAT@nmfmt{\NAT@nm}%
     \hyper@natlinkbreak{\NAT@spacechar\NAT@@open\if*#1*\else#1\NAT@spacechar\fi}%
       {\@citeb\@extra@b@citeb}%
     \NAT@date\NAT@@close}}
  {}{\PackageWarning{manuscript-main}{citet close-paren patch FAILED}}
\patchcmd{\NAT@citex}
  {\ifNAT@swa\else\if*#2*\else\NAT@cmt#2\fi
     \if\relax\NAT@date\relax\else\NAT@@close\fi\fi}
  {\ifNAT@swa\else\if*#2*\else\NAT@cmt#2\fi\fi}
  {}{\PackageWarning{manuscript-main}{citet trailing-close patch FAILED}}
\makeatother

\usepackage{indentfirst}
\makeatletter
\renewcommand{\@maketitle}{%
  \vbox{%
    \hsize\textwidth
    \linewidth\hsize
    \vskip 0.1in
    \raggedright
    \hyphenpenalty=10000\exhyphenpenalty=10000
    {\fontsize{17}{20}\selectfont\bfseries \@title\par}
    \vskip 0.2in
    {\raggedright \@author \par}
    \vskip 0.4in
  }%
}
\makeatother

\newcommand{\affmark}[1]{\textsuperscript{#1}}

\title{Partial Inverse Design of High-Performance Concrete Using Cooperative Neural Networks for Constraint-Aware Mix Generation
}

\author{%
  {\normalsize\bfseries
    Agung Nugraha\affmark{1},
    Heungjun Im\affmark{1},
    Jihwan Lee\affmark{1}\thanks{\textit{Corresponding author:} \texttt{jihwan@pknu.ac.kr}}%
  }\\[0.1in]
  {\small
    \affmark{1}\,Department of Industrial and Data Engineering, Major in Industrial Data Science and Engineering, Pukyong National University, Busan, 48513, Republic of Korea
  }%
}

\begin{document}
\maketitle
\nolinenumbers

\singlespacing

\input{manuscript-sec-abstract}

\input{manuscript-sec-introduction}

\input{manuscript-sec-methodology}

\input{manuscript-sec-experimental_setup}

\input{manuscript-sec-results}

\input{manuscript-sec-conclusion}

\input{manuscript-sec-acknowledgements}

\input{manuscript-sec-appendix}

%Bibliography
\bibliographystyle{reference}
\bibliography{references}

\end{document}

%% file: manuscript-sec-abstract.tex
\begin{abstract}
High-performance concrete (HPC) requires complex mix design decisions involving interdependent variables and practical constraints. While data-driven methods have improved predictive modeling for forward design in concrete engineering, inverse design remains limited, especially when some variables are fixed and only the remaining ones must be inferred. This study proposes a cooperative neural network framework for the partial inverse design of HPC. The framework integrates an imputation model with a surrogate strength predictor and learns through cooperative training. Once trained, it generates valid and performance-consistent mix designs in a single forward pass without retraining for different constraint scenarios. Compared with baseline models, including autoencoder models and Bayesian inference with Gaussian process surrogates, the proposed method achieves strength consistency between the surrogate-predicted strength of the generated mixes and the target strength with R-squared values of 0.84 to 0.89 and substantially reduces the mean squared error of this strength consistency by approximately 42\% and 60\%, respectively. The results demonstrate a novel, accurate, and computationally efficient application of artificial intelligence in concrete science by applying a cooperative neural network for constraint-aware partial inverse design of HPC mix generation.
\end{abstract}

% keywords can be removed
\keywords{High-performance concrete\and Data-driven design\and Partial inverse design\and Surrogate model\and Imputation model\and Autoencoder}

%% file: manuscript-sec-introduction.tex
\section{Introduction}
\label{sec:introduction}

    High-performance concrete (HPC) is an advanced construction material valued for its superior strength and durability. Unlike conventional concrete, which primarily consists of Portland cement, fine and coarse aggregates, and water, HPC incorporates supplementary cementitious materials (SCMs) such as fly ash and blast furnace slag, together with chemical admixtures such as superplasticizers \citep{ke_bayesian_2021,yeh_modeling_1998}. These blended cement systems enhance workability, improve long-term performance, and reduce the environmental impact by lowering the dependence on Portland cement, which is a significant contributor to CO$_2$ emissions. As a result, HPC supports both structural efficiency and environmental sustainability in construction \citep{ke_bayesian_2021,aitcin_durability_2003}.

    Alongside material innovations and the increasing availability of high-quality experimental data, the data-driven design (DDD) paradigm has emerged as a transformative approach in concrete research and materials science \citep{fazel_zarandi_fuzzy_2008,wijesundara_machine_2025,challapalli_inverse_2021,wanigasekara_machine_2021}. DDD is broadly defined as a methodology in which empirical or simulated data serve as the foundation for predictive or generative modeling to support engineering decision-making \citep{wang_implications_2022,jiang_data-driven_2022}. Unlike conventional approaches that rely heavily on explicit physical or mechanistic models, DDD employs predictive models that capture the complex and often nonlinear relationships directly from data. In doing so, it enables rapid property prediction, reduces the need for time-consuming experiments or costly simulations, and facilitates the intelligent exploration of complex design spaces.

    \begin{figure}[htbp]
        \centering
        \includegraphics[width=1.0\textwidth]{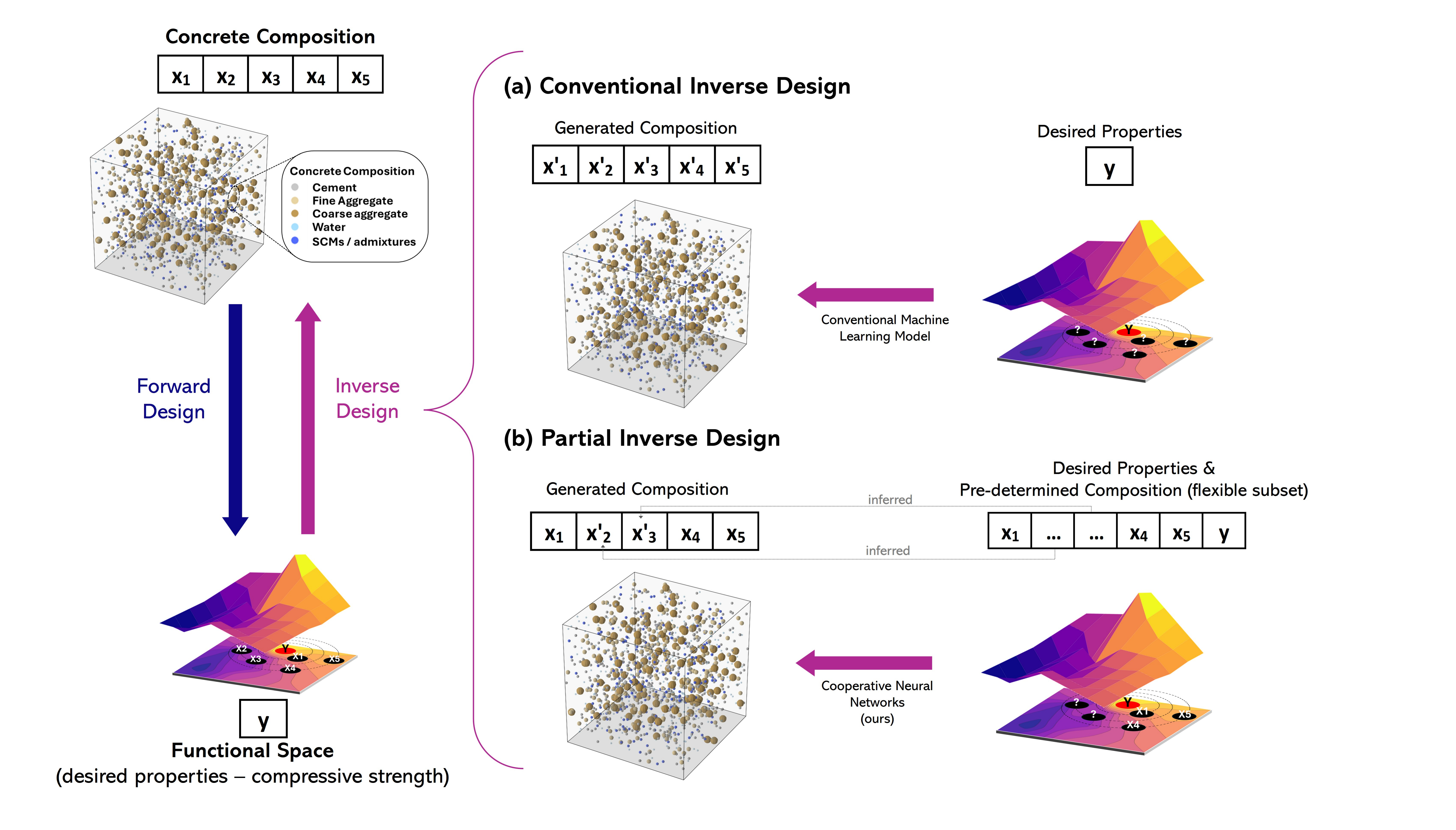}
        \caption{The DDD paradigm of forward design, inverse design and partial inverse design.}
        \label{fig:myfigure1}
    \end{figure}

    Within the data-driven design paradigm, design tasks can be categorized into distinct approaches. Forward design (as illustrated in \figref{fig:myfigure1}) represents the predictive approach, in which known input variables (e.g., mix proportions) are used to estimate material properties such as compressive strength. This process involves developing a predictive or surrogate model that serves as an efficient substitute for complex and computationally expensive physical models \citep{ebrahimian_material_2024,miao_surrogate_2023}. The surrogate model learns from available data on design parameters and their corresponding performance metrics, enabling rapid and reliable evaluation of candidate mix design solutions. One of the early applications of machine learning-based surrogate modeling in the concrete domain was introduced by \cite{yeh_modeling_1998}, who developed an artificial neural network (ANN) model to predict compressive strength of HPC. The study concluded that ANN outperformed traditional regression models in accuracy and proved to be a convenient tool for analyzing the effects of mix proportion variables through numerical experiments. Following this methodology, \cite{yeh_analysis_2006} also applied neural networks as surrogate models in combination with design of experiments to investigate the effects of fly ash replacement, curing age, and water-binder ratio on the compressive strength of both low- and high-strength concrete. Later on, with the rapid advancement of machine learning research, various algorithms were applied in the cement and concrete domain \citep{li_machine_2022}, including ANN \citep{xu_parametric_2019, ouyang_predicting_2020,nguyen_efficient_2021,hossain_regression_2018,nguyen-sy_predicting_2020}, polynomial neural network hybrid models \citep{fazel_zarandi_fuzzy_2008}, support vector machine (SVM) and its hybrid model \citep{nguyen_efficient_2021, nguyen-sy_predicting_2020,cheng_high-performance_2012}, Gaussian process (GP) model \citep{ke_bayesian_2021,zhang_multiple_2019}, decision tree-based model \citep{zhang_multiple_2019, zhang_toward_2020}, ensemble model \citep{wijesundara_machine_2025,ouyang_predicting_2020,nguyen_efficient_2021,nguyen-sy_predicting_2020, zhang_multiple_2019, zhang_toward_2020,oey_machine_2020,marani_machine_2020}, and other combination or hybrid approaches \citep{chou_concrete_2012,m_optimized_2025}.  \cite{fazel_zarandi_fuzzy_2008} constructed and evaluated six hybrid neural network frameworks integrating fuzzy logic with polynomial neural networks, aiming to identify the most effective model for predicting the compressive strength of concrete mix designs. \cite{cheng_high-performance_2012} developed a hybrid model that integrated Fuzzy Logic (FL), weighted SVM, and fast messy genetic algorithm (fmGA) into an adaptive system called Evolutionary Fuzzy Support Vector Machine Inference Model for Time Series Data (EFSIMT), demonstrating its effectiveness in predicting the compressive strength of concrete. \cite{ke_bayesian_2021} employed GP models as surrogate models in their inverse design architecture to predict concrete compressive strength, highlighting their ease of integration with Bayesian inference methods. More recently, the application of surrogate models to predict compressive strength in advanced concretes has been demonstrated by several studies. \cite{wijesundara_machine_2025} evaluated seven predictive models, including SVM, ensemble tree models, and multi-layer perceptron (MLP), for ultra-high-performance fiber-reinforced concrete (UHPFRC). \cite{m_optimized_2025} developed a hybrid model that combines ANN with a genetic algorithm (GA), referred to as ANN-GA, for HPC.

    In contrast, inverse design represents a target-driven paradigm that reverses the usual direction of prediction. Rather than estimating material properties from known mix compositions, it determines the combination of design variables required to achieve a specified target property, such as compressive strength (\figref{fig:myfigure1}). Conceptually, it seeks to map desired outcomes back to feasible material configurations, a process often referred to as design inversion in materials informatics. In data-driven contexts, inverse design architectures are generally realized in two forms: (a) single-stage approaches, where the configurations are determined directly using models such as generative neural networks, and (b) two-stage approaches, where surrogate models are integrated with complementary mechanisms to refine or guide the generation of valid designs \citep{nugraha_cooperative_2025}. In concrete research, these mechanisms have led to several methodological approaches, including optimization-based, Bayesian and probabilistic inference, and generative approaches, each differing in how the inversion process is formulated.

    The most established class in concrete research comprises optimization-based approaches, where a surrogate predictive model is first trained to approximate the relationship between design variables and material properties, and then combined with a search-based method or an optimization algorithm to iteratively identify compositions satisfying the target performance. \cite{huang_intelligent_2020} developed an inverse design framework that coupled SVR with a Firefly Algorithm to optimize mixture proportions for steel fiber–reinforced concrete (SFRC). Similarly, \cite{zhang_hybrid_2020} proposed a hybrid intelligent system that integrates multiple AI-based surrogate models with a metaheuristic optimizer to design recycled aggregate concrete satisfying multiple objectives. Recently, \cite{le_nguyen_generative_2024} employed several predictive tree-based machine learning models coupled with a Fast Elitist Non-Dominated Sorting Genetic Algorithm for Multi-Objective Optimization (NSGA-II) to iteratively evolve optimal concrete mix proportions. Utilizing similar machine learning model, \cite{zhang_transfer_2025} introduced a Bayesian Optimization XGBoost (BO-XGBoost) framework for lightweight strain-hardening UHPC, where transfer learning was applied to the XGBoost surrogate to enhance generalization under data scarcity. These surrogate-guided optimization methods enable interpretable and flexible design exploration but can be computationally intensive due to their iterative nature.

    Beyond optimization-based frameworks, other approaches have emerged that tackle the inverse problem from different perspectives. Bayesian and probabilistic inference methods treat inverse design as a problem of posterior estimation rather than optimization, inferring a distribution of feasible inputs that satisfy a target property and thereby capturing uncertainty in the design space. \cite{ke_bayesian_2021} exemplified this paradigm by combining a GP surrogate with the Metropolis–Hastings (MH) algorithm in a Markov Chain Monte Carlo (MCMC) framework, enabling probabilistic inference of high-performance concrete compositions that achieve desired compressive strengths. In parallel, direct-generation approaches based on generative or encoder-decoder neural networks have gained attention for learning the inverse mapping explicitly, thereby eliminating the need for iterative search. Once trained, these models can generate candidate mix designs in a single forward pass conditioned on desired performance targets. \cite{yu_generative_2023}, for example, developed a generative AI framework for the performance-based design of engineered cementitious composites (ECC), employing an invertible neural network (INN) to directly generate mix proportions that satisfy target tensile strength and strain requirements. While Bayesian approaches emphasize probabilistic reasoning and uncertainty estimation, direct-generation models focus on efficient one-shot inference. Despite their respective computational and data challenges, both offer promising directions for advancing data-driven inverse design in concrete mix design.

    Despite recent advances, existing inverse design studies share a fundamental limitation that they are primarily designed to generate complete mix designs and cannot flexibly handle situations where only part of the composition should be optimized \citep{ke_bayesian_2021, huang_intelligent_2020, zhang_hybrid_2020, le_nguyen_generative_2024, zhang_transfer_2025, yu_generative_2023}. Most frameworks assume that all design variables are freely adjustable, but in practice, this assumption may not always hold. Concrete mix design in real construction projects is subject to multiple constraints, where certain variables are predetermined by construction standards, cost limits, or material availability. In these cases, designers often need to adjust only a subset of variables while keeping others fixed to satisfy project-specific or regulatory requirements \citep{noauthor_eurocode_2004}. However, existing inverse design methods cannot selectively infer unknown variables while maintaining fixed ones, making them unsuitable for modular or hierarchical design workflows where partial reconfiguration of a mix is required instead of complete redesign.

    This challenge defines the partial inverse design problem, in which only a portion of design variables is unknown and must be estimated under specific constraints. Conceptually, this problem can be formulated as a constrained optimization task \citep{mathern_multi-objective_2021}. Prior approaches have typically relied on iterative optimization frameworks, such as Bayesian optimization or other search-based methods \citep{wang_constrained_2024, amini_constrained_2025}, which are computationally expensive and difficult to generalize to varying constraint conditions.

    To address this limitation, this study builds on the Cooperative Neural Network (CoNN) framework of \citet{nugraha_cooperative_2025} and applies it to the partial inverse design of HPC. The framework reformulates partial inverse design as a constraint-aware imputation problem, in which missing design variables are reconstructed to meet a specified target property. It integrates two key components, an imputation model based on an autoencoder architecture that reconstructs incomplete mix designs from partially masked inputs, and a surrogate performance predictor that estimates compressive strength from complete designs and provides feedback during cooperative training. During training, the two models interact in a cooperative learning loop, where the imputation model generates candidate designs for missing variables and the surrogate model evaluates their predicted strength to guide reconstruction accuracy. This joint optimization enables the network to produce designs that are both numerically consistent with the training data and physically plausible with respect to target performance. Unlike iterative optimization methods such as Bayesian optimization, the trained CoNN generates feasible mix compositions in a single forward pass, allowing instant inference without retraining even when constraint conditions change. The framework is evaluated on a benchmark HPC dataset to validate its accuracy, stability, and efficiency under limited-data conditions. Relative to \citet{nugraha_cooperative_2025}, this study contributes in three ways. First, at the paradigm level, it positions partial inverse design as a practical constraint-aware generation approach for HPC under data-scarcity conditions. Second, at the methodological level, it adds empirical min--max feature clipping as an implicit plausibility safeguard for CoNN, and it extends the Bayesian inverse-design framework which originally developed for conventional inverse design of HPC to the partial inverse design setting for head-to-head comparison. Third, at the empirical level, it augments the evaluation with paired $t$-test and Wilcoxon signed-rank significance testing and presents a scenario-specific demonstration of operational use on HPC mix design. The results show that the CoNN achieves higher predictive accuracy and computational efficiency than conventional search-based methods, demonstrating its suitability for data-driven material design tasks where only part of the composition is adjustable.

    The remainder of this paper is structured as follows. \secref{sec:method} details the methodology, including the components and integration of the cooperative framework for inverse design of HPC. \secref{sec:experiment} outlines the experimental and evaluation setup, including dataset description, preprocessing, training and inference configurations, evaluation objective, and baseline models for comparison. \secref{sec:results} presents the experimental results, highlighting key findings and practical implications. Finally, \secref{sec:conclusion} concludes the paper by summarizing the contributions and discussing limitations and directions for future research.

%% file: manuscript-sec-methodology.tex
\section{Cooperative Neural Networks}
\label{sec:method}

    This study adapts the Cooperative Neural Network (CoNN) framework to the problem of partial inverse design of HPC under multi-constraint conditions. In this context, partial inverse design refers to estimating the unknown portion of a mix composition when some ingredients are predetermined by engineering constraints or experimental setup, while still satisfying a target compressive strength. Formally, each mix-design instance $\mathbf{x} \in \mathbb{R}^{d}$ is partitioned into $\mathbf{x}_{\mathrm{free}}$, the subset of variables to be inferred, and $\mathbf{x}_{\mathrm{fixed}}$, the subset held fixed by the constraint pattern. Given a target performance $y_{\mathrm{target}}$, the inverse mapping is realized either through a generative form $p_{\theta}(\mathbf{x}_{\mathrm{free}} \mid y_{\mathrm{target}}, \mathbf{x}_{\mathrm{fixed}})$ or a non-generative form $\hat{\mathbf{x}}_{\mathrm{free}} = h_{\theta}(y_{\mathrm{target}}, \mathbf{x}_{\mathrm{fixed}})$ \citep{nugraha_cooperative_2025}. The CoNN architecture couples two models, a surrogate model and an imputation model that learn cooperatively to reconstruct feasible mix designs from incomplete inputs.

\begin{figure}[htbp]
        \centering
        \includegraphics[width=1.0\textwidth]{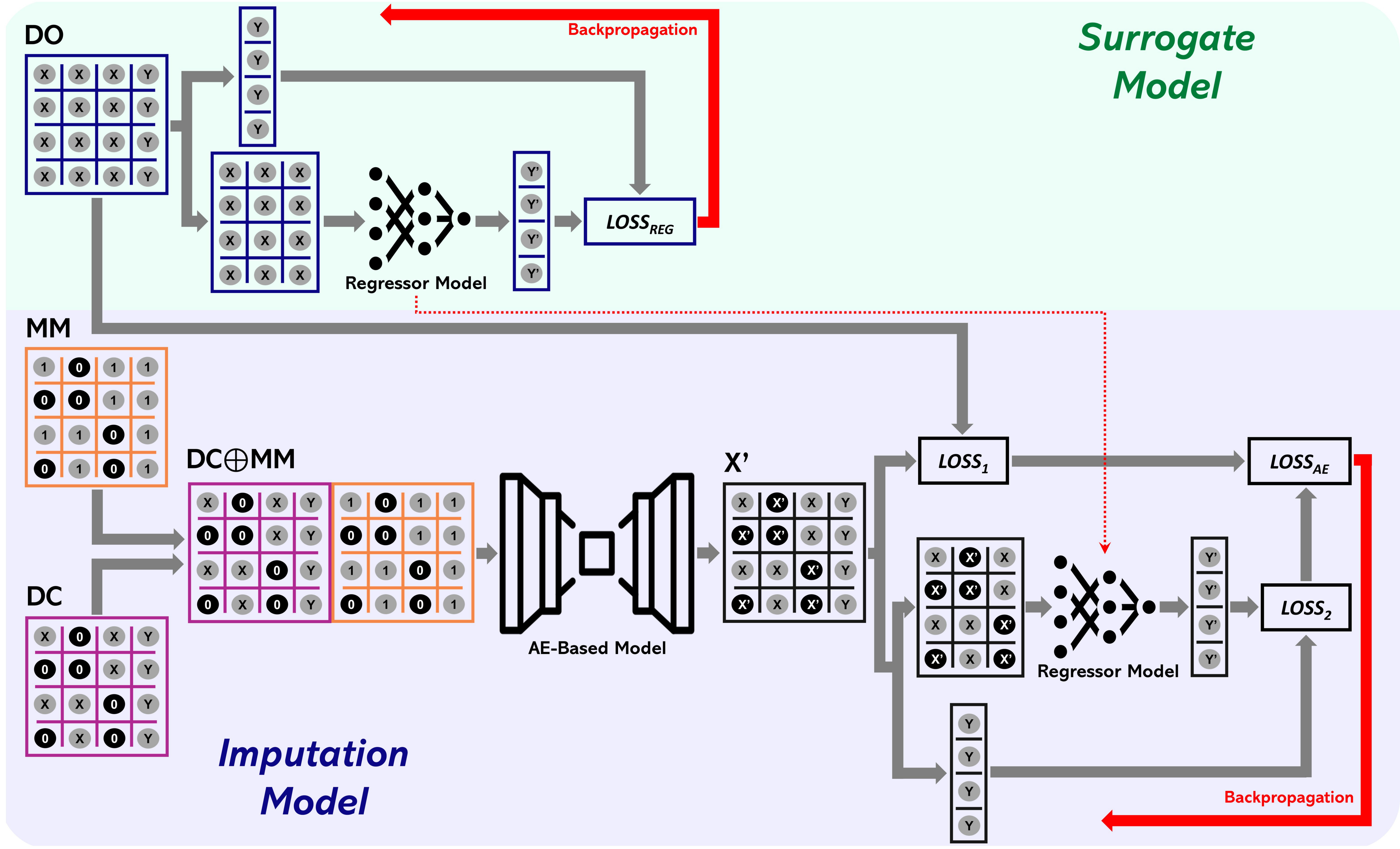}
        \caption{Cooperative Neural Network (CoNN) (adopted from \cite{nugraha_cooperative_2025}). Schematic illustration of the CoNN framework for partial inverse design of HPC. The upper (green) region shows the surrogate model predicting compressive strength from complete mix designs, while the lower (purple) region shows the autoencoder-based imputation model reconstructing masked variables. CoNN is optimized cooperatively through reconstruction ($\mathcal{L}_{1}$) and performance ($\mathcal{L}_{2}$) objectives, combined into a unified loss ($\mathcal{L}_{\mathrm{AE}}$). Red arrows indicate the backpropagation flow connecting both models, forming a feedback loop that aligns reconstructed compositions with desired performance targets.}
        \label{fig:myfigure2}
    \end{figure}

    The framework operates on a dataset $\mathbf{X} \in \mathbb{R}^{n \times d}$, where each row corresponds to a concrete batch (either experimentally measured or simulated) and each column represents one design variable such as cement, water, blast-furnace slag (BFS), fly ash (PFA), superplasticizer (SP), fine aggregate (FA), coarse aggregate (CA), and curing age. The corresponding target property $\mathbf{y} \in \mathbb{R}^{n}$ denotes the experimentally measured compressive strength (MPa). In practice, this dataset can be obtained from laboratory experiments, field tests, or simulation-based mix-design studies, depending on availability. The overall architecture of the proposed framework is illustrated in \figref{fig:myfigure2}.

    \subsection{Surrogate Model}
        The surrogate model, depicted in the upper part of \figref{fig:myfigure2}, approximates the nonlinear mapping between mix composition and compressive strength as $\hat{\mathbf{y}} = f(\mathbf{X})$, where $f(\cdot)$ denotes an ANN regressor. From the original dataset $\mathbf{DO}$, the model takes the complete mix-design variables $\mathbf{X}$ as input and predicts the corresponding compressive strength $\hat{\mathbf{y}}$. Training minimizes the mean-squared error between the measured and predicted strengths:

        \begin{equation}
        \mathcal{L}_{\mathrm{REG}} = \frac{1}{N} \sum_{i=1}^{N} \left( y_i - \hat{y}_i \right)^2
        \label{eq:eq1}
        \end{equation}

        where $N$ is the number of samples. Minimizing $\mathcal{L}_{\mathrm{REG}}$ enables the surrogate model to capture the forward relationship between concrete mix proportions and compressive strength. Once trained, it collaborates with the imputation model, providing performance-based feedback that guides the generation of strength-consistent mix designs.

    \subsection{The Imputation Model}
        The imputation model, shown in the lower part of \figref{fig:myfigure2}, is based on an autoencoder (AE) architecture that reconstructs missing mix variables from partially specified data. By learning a compact latent representation, the model infers unknown variables from the available context. Its inputs are:
        \begin{itemize}
            \item $\mathbf{DO}$: the original complete mix-design data (including the performance criterion $\mathbf{y}$),
            \item $\mathbf{DC}$: a corrupted version of $\mathbf{DO}$ obtained by replacing the $\mathbf{x}_{\mathrm{free}}$ positions of $\mathbf{X}$ with zeros while the $\mathbf{y}$ entry is retained, and
            \item $\mathbf{MM}$: a binary mask matrix indicating observed ($1$) and missing ($0$) variables.
        \end{itemize}
        To make the encoder aware of which features are missing, $\mathbf{DC}$ and $\mathbf{MM}$ are concatenated column-wise to form the input $(\mathbf{DC}\oplus \mathbf{MM})$. The encoder $f(\cdot)$ transforms this combined input into a latent representation.

        \begin{equation}
        \mathbf{z} = f(\mathbf{DC} \oplus \mathbf{MM})
        \label{eq:eq2}
        \end{equation}

        The decoder $g(\mathbf{z})$ reconstructs a complete mix design. To ensure that the generated values remain within the empirically valid domain (e.g., unrealistic or physically infeasible mix quantities), feature clipping is applied by constraining each reconstructed variable to the per-feature minimum ($\mathbf{X}_{\min}$) and maximum ($\mathbf{X}_{\max}$) values from the training dataset:

        \begin{equation}
        \hat{\mathbf{X}} = \min \left( \max \left( g(\mathbf{z}), \mathbf{X}_{\min} \right), \mathbf{X}_{\max} \right)
        \label{eq:eq3}
        \end{equation}

        An ablation study that isolates the effect of this clipping step on two concrete-specific compound ratios, namely the water-to-binder ratio and the supplementary cementitious replacement ratio, is reported in \appref{sec:appendix_clipping}.

        The reconstructed output is merged with the unmasked components of the input to obtain the completed design parameters

        \begin{equation}
        \mathbf{X}' = (\mathbf{1} - \mathbf{MM}) \odot \hat{\mathbf{X}} + \mathbf{DC}
        \label{eq:eq4}
        \end{equation}

        where $\odot$ denotes element-wise multiplication.

    \subsection{Cooperative Learning Mechanism}
        During training, the imputation model generates completed design parameters $\mathbf{X}'$, and the surrogate model evaluates their predicted compressive strength as $\hat{\mathbf{y}} = f(\mathbf{X}')$. Two complementary objectives are optimized jointly: the reconstruction loss ($\mathcal{L}_{1}$), which measures the difference between the reconstructed and original designs, and the performance loss ($\mathcal{L}_{2}$), which quantifies the deviation between predicted and measured strengths.

        The reconstruction loss $\mathcal{L}_{1}$ is formulated as the mean squared error between the completed design $\mathbf{X}'$ and the original uncorrupted dataset $\mathbf{DO}$ across $N$ training samples.

        \begin{equation}
        \mathcal{L}_{1} = \frac{1}{N} \sum_{i=1}^{N} \left( \mathbf{DO}_{i} - \mathbf{X}'_{i} \right)^{2}
        \label{eq:L1}
        \end{equation}

        The performance loss $\mathcal{L}_{2}$ is formulated as the mean squared error between the measured strengths $\mathbf{y}$ and the predicted strengths $\hat{\mathbf{y}}$ across $N$ training samples.

        \begin{equation}
        \mathcal{L}_{2} = \frac{1}{N} \sum_{i=1}^{N} \left( y_{i} - \hat{y}_{i} \right)^{2}
        \label{eq:L2}
        \end{equation}

        These are combined into a unified cooperative loss.

        \begin{equation}
        \mathcal{L}_{\mathrm{TOTAL}} = \mathcal{L}_{\mathrm{AE}} = \alpha \, \mathcal{L}_{1} + (1 - \alpha) \, \mathcal{L}_{2}
        \label{eq:eq5}
        \end{equation}

        where $\alpha \in [0,1]$ controls the trade-off between reconstruction fidelity and performance consistency. Through this cooperative feedback, the imputation model learns to generate statistically plausible and performance-compliant compositions. This optimization establishes the core mechanism of the CoNN.

    \subsection{Imputation Model Variants}
        Depending on the target application, the imputation model can adopt deterministic or generative denoising variants while maintaining the same cooperative structure:
        \begin{itemize}
            \item \textbf{Cooperative DAE (Denoising Autoencoder):} a deterministic model minimizing $\mathcal{L}_{\mathrm{TOTAL}}$, suited for one-to-one reconstruction of specific HPC mix designs.

            \item \textbf{Cooperative DVAE (Denoising Variational Autoencoder):} introduces a probabilistic latent variable and a Kullback--Leibler divergence term $D_{\mathrm{KL}}\!\left(q(\mathbf{z} \mid \mathbf{X}) \,\|\, p(\mathbf{z})\right)$ to capture uncertainty and generate multiple valid design candidates.

            \item \textbf{Cooperative DWAE (Denoising Wasserstein Autoencoder):} replaces the KL divergence with a Maximum Mean Discrepancy (MMD) term $\mathrm{MMD}_{k}^{2}\!\left(q(\mathbf{z}), p(\mathbf{z})\right)$ to enforce smoother latent distributions and improve robustness under limited data.
        \end{itemize}
        All variants share the same cooperative loss defined in \eqnref{eq:eq5}, each with slight modifications reflecting their respective objective functions, and the same two-model structure shown in \figref{fig:myfigure2}. This flexibility allows the framework to support both deterministic optimization and generative exploration, enabling the generation of diverse, performance-consistent mix designs within a unified learning architecture. More details on the VAE variant can be found in the original work by \cite{kingma_auto-encoding_2013}, and on the WAE variant in the work by \cite{tolstikhin_wasserstein_2019}.

%% file: manuscript-sec-experimental_setup.tex
\section{Experimental Setup and Evaluation}
\label{sec:experiment}
    This section describes the dataset, preprocessing, and experimental configuration used to evaluate the CoNN framework. The setup was designed to ensure a fair and comprehensive assessment of model performance and to provide consistent conditions for comparison with baseline approaches.

    \begin{table}[htbp]
        \centering
        \caption{Dataset information of HPC.}
        \label{tab:tab1}
        \begin{tabular}{l l}
        \hline
        \textbf{Feature Name} & \textbf{Description (with Units)} \\
        \hline
        Cement   & Cement content in the mix (kg/m$^3$) \\
        BFS      & Blast Furnace Slag content (kg/m$^3$) \\
        PFA      & Fly Ash content (kg/m$^3$) \\
        Water    & Amount of water in the mix (kg/m$^3$) \\
        SP       & Superplasticizer dosage (kg/m$^3$) \\
        CA       & Coarse Aggregate content (kg/m$^3$) \\
        FA       & Fine Aggregate content (kg/m$^3$) \\
        Age      & Concrete curing time (days) \\
        Strength & Compressive strength of concrete (MPa) \\
        \hline
        \end{tabular}
    \end{table}

    \subsection{Dataset Description}
        A publicly available concrete dataset originally collected by \cite{yeh_modeling_1998} was used for all experiments. This dataset has been widely adopted in previous data-driven concrete research and thus provides a reliable benchmark for comparative studies \mbox{~\citep{chou_concrete_2012, cheng_high-performance_2012, ke_bayesian_2021, m_optimized_2025}}. It contains 1,030 records of concrete mix ingredients and their corresponding compressive strengths. The variables and their physical units are summarized in \tabref{tab:tab1}.

        To align with the focus on HPC, only samples with curing ages of up to 28 days were retained, as this period captures the majority of strength development in HPC \citep{ke_bayesian_2021, zhang_transfer_2025, noauthor_eurocode_2004}. Furthermore, this restriction was also applied to emphasize early-age strength development, representing the most critical curing period for HPC \citep{aitcin_durability_2003}. After filtering, a total of 749 records out of 1,030 concrete specimens were used in this study.

    \subsection{Data Preprocessing and Splitting}
        \label{sec:datasplit}
\begin{figure}[htbp]
            \centering
            \includegraphics[width=1.0\textwidth]{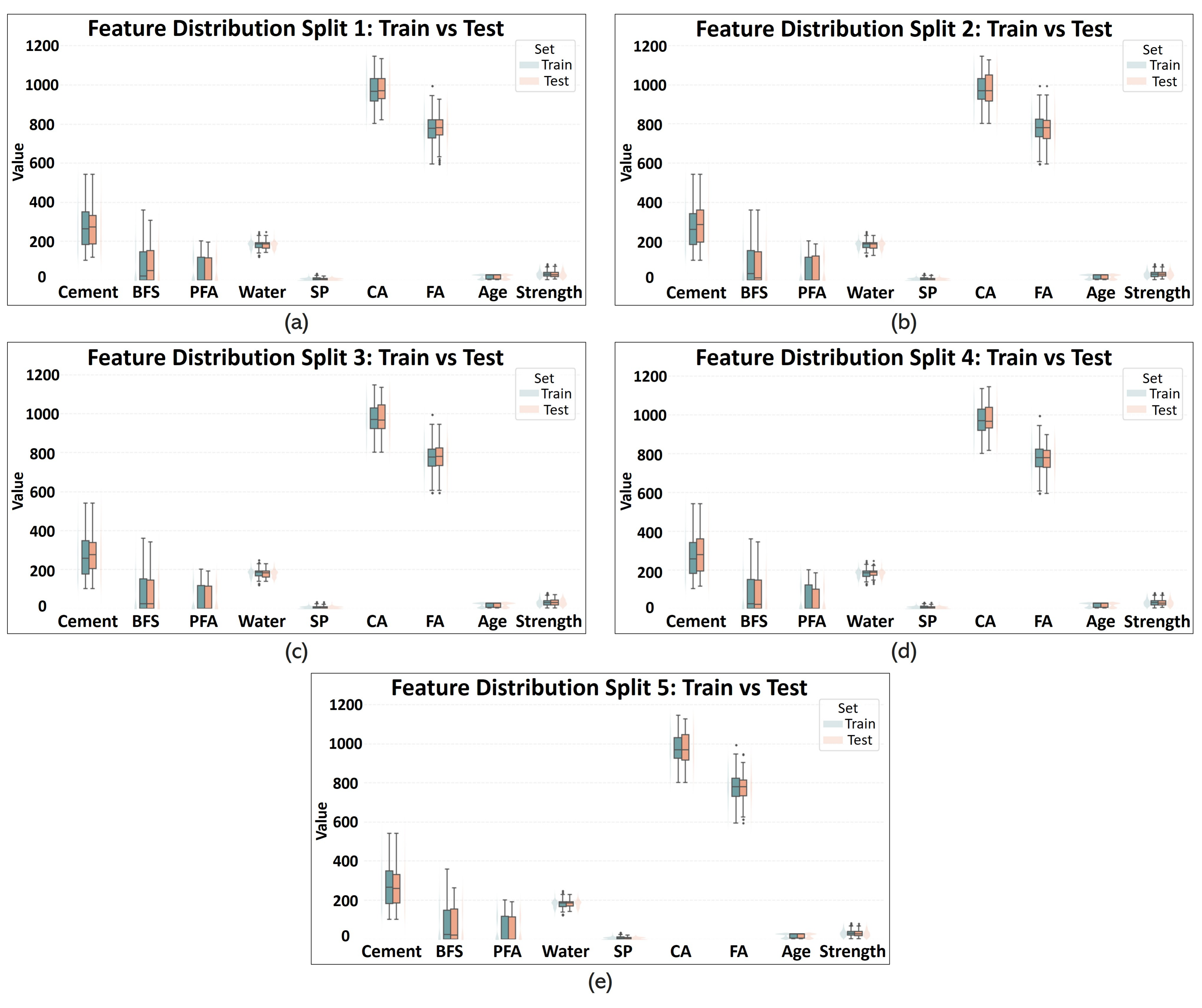}
            \caption{Data distribution of the five-split experiment setup (original scale).}
            \label{fig:myfigure3}
        \end{figure}

        The dataset was divided into training and testing subsets using an 8:2 split ratio, which balanced model learning with unbiased generalization assessment. Because the dataset is relatively small, we repeated the splitting process five times with different random seeds and reported the average performance across these five-seed repeated hold-out evaluation. Each split produced 600 training and 149 testing samples. \figref{fig:myfigure3} confirms that no noticeable distributional bias was introduced by the splitting process. All input variables were normalized so that features measured on different scales contributed comparably during training and no single variable dominated the gradient updates \citep{ke_bayesian_2021}. The StandardScaler is refit on the training partition within each of the five random splits, and the resulting training-set statistics are used to transform the test partition, precluding any test-partition leakage.

    \subsection{Model Training and Inference Procedure}
        \label{sec:training}
        During training, the CoNN framework jointly optimizes the surrogate and imputation models using the cooperative objective defined in \eqnref{eq:eq5}, with the balancing coefficient set to $\alpha = 0.5$ to weight the reconstruction and performance losses equally, following the convention adopted in the original CoNN framework of \citet{nugraha_cooperative_2025}. The surrogate model learns to predict compressive strength from complete mix designs, while the imputation model reconstructs masked inputs based on known variables and the performance feedback provided by the surrogate.

        The network architectures of the surrogate model and the imputation autoencoder are summarized in \tabref{tab:network_arch}. The imputation encoder receives an 18-dimensional input formed by concatenating the 9-dimensional design vector (eight mix variables and one target strength) with the 9-dimensional mask vector, and its decoder produces a 9-dimensional reconstruction that is merged with the original design through the mask matrix as in \eqnref{eq:eq4}. For the variational and Wasserstein variants (DVAE and DWAE), the final encoder layer is duplicated into two parallel heads of the same dimensions ($16 \to 8$) that produce the latent mean and variance parameters, while all remaining layers are identical across the three imputation variants. The corresponding training hyperparameters are reported in \tabref{tab:training_hyperparams}. Both networks are trained from scratch independently within each random data split following the repeated hold-out evaluation protocol described in \secref{sec:datasplit}, with the loss formulations for the surrogate and imputation networks following the definitions introduced in \secref{sec:method}.

        \begin{table}[htbp]
            \centering
            \caption{Network architectures of the surrogate and imputation autoencoder used in the CoNN framework.}
            \label{tab:network_arch}
            \renewcommand{\arraystretch}{1.25}
            \setlength{\tabcolsep}{6pt}
            \begin{tabular}{lcc}
            \toprule
            \textbf{Network} & \textbf{Layer dimensions} & \textbf{Activation} \\
            \midrule
            Surrogate ANN       & $8 \to 60 \to 48 \to 32 \to 24 \to 12 \to 1$         & ReLU \\
            Imputation Encoder  & $18 \to 512 \to 256 \to 128 \to 64 \to 16 \to 8$     & ReLU \\
            Imputation Decoder  & $8 \to 16 \to 64 \to 128 \to 256 \to 512 \to 9$      & ReLU \\
            \bottomrule
            \end{tabular}
        \end{table}

        \begin{table}[htbp]
            \centering
            \caption{Training hyperparameters for the surrogate and imputation networks.}
            \label{tab:training_hyperparams}
            \renewcommand{\arraystretch}{1.25}
            \setlength{\tabcolsep}{6pt}
            \begin{tabular}{lcc}
            \toprule
            \textbf{Hyperparameter} & \textbf{Surrogate} & \textbf{Imputation} \\
            \midrule
            Optimizer               & Adam                & Adam                \\
            Learning rate & $1\times10^{-3}$    & $5\times10^{-4}$    \\
            Batch size              & 16                  & 16                  \\
            Epochs                  & 150                 & 150                 \\
            Dropout rate (hidden layers) & 0.1 & -- \\
            \bottomrule
            \end{tabular}
        \end{table}

        In the inference phase, the trained model receives partial mix specifications, where some variables are fixed by the user or by engineering constraints and others are left unspecified. A binary mask matrix encodes which variables are known (1) and which must be inferred (0). The masked design, along with the desired target strength, is passed through the imputation model to generate candidate values for the missing variables.

\begin{figure}[htbp]
            \centering
            \includegraphics[width=1.0\textwidth]{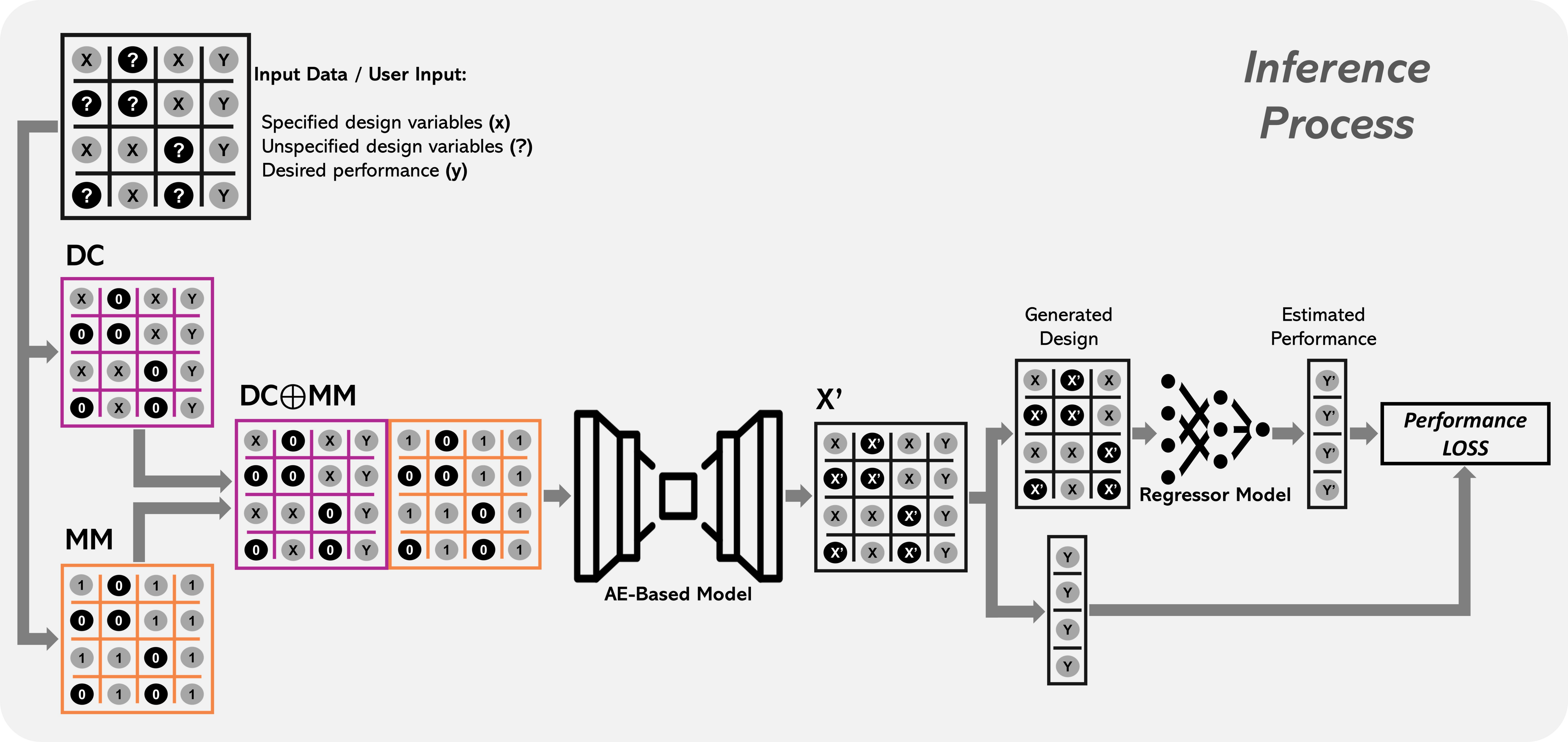}
            \caption{Inference process of Cooperative Neural Network (CoNN) (adopted from \cite{nugraha_cooperative_2025}).}
            \label{fig:myfigure4}
        \end{figure}

        The same masking protocol is applied to both the training and testing partitions, so that the imputation model is exposed during training to the same partial-specification inputs it must reconstruct at inference. For every sample, the number of masked input variables is drawn uniformly between one and five and the masked positions are sampled uniformly across the mix-variable columns, with the curing-age variable (Age) and the performance target $y_{\mathrm{target}}$ always retained. Reported metrics are averaged across this random distribution of masking levels, with larger counts progressively increasing the difficulty of inference. All other experimental conditions, including network architecture, training hyperparameters, and the positions of masked variables were held constant across all experiments to ensure fair comparison. Because the imputation model explicitly receives the mask matrix as part of its input, it helps as a hint which variables are fixed and which are missing. This design enables the framework to adapt to different constraint scenarios without retraining, allowing multiple partial design configurations to be inferred simply by changing the mask input in a single forward pass. The overall inference process is illustrated in \figref{fig:myfigure4}.

    \subsection{Evaluation Objectives}
        \label{sec:eval_objective}
        Model performance was evaluated with two complementary objectives, prediction accuracy and computational efficiency. Prediction accuracy is defined as how closely the predicted performance of generated designs matched the target performance (compressive strength). For this, we measure the coefficient of determination ($R^2$), mean absolute error (MAE), and mean squared error (MSE), averaged over the five evaluations. Their definitions are summarized in \tabref{tab:tab2}, with values reported on the normalized strength scale. Meanwhile, computational efficiency is evaluated using the inference time. This is the measured wall-clock time needed to generate mix designs for the full test partition.

        Beyond mean accuracy, we assessed whether the differences between methods were statistically significant using paired tests on the per-condition metric values. Combining the five random splits in \secref{sec:datasplit} with the five masking configurations in \secref{sec:training} yielded $n = 25$ matched pairs per metric, where each pair fixed the split and masking setting and varied only the training scheme. We applied the paired $t$-test and the paired Wilcoxon signed-rank test, using a sign convention in which a positive mean difference indicates that the cooperative framework outperforms the baseline. A $p$-value below $0.05$ was considered statistically significant, and reporting both tests helped confirm that the conclusions were not sensitive to the normality assumption of the $t$-test.

        \begin{table}[htbp]
            \centering
            \caption{Prediction accuracy metrics.}
            \label{tab:tab2}
            \renewcommand{\arraystretch}{1.5} % increases row height
            \begin{tabular}{>{\raggedright\arraybackslash}p{7cm} c}
            \toprule
            \textbf{Metrics} & \textbf{Definition} \\
            \midrule
            Coefficient of Determination ($R^2$)
            & $\displaystyle
            1 - \frac{\sum_{i=1}^{m} (y_i - \hat{y}_i)^2}
            {\sum_{i=1}^{m} (y_i - \bar{y})^2}
            $ \\

            Mean Absolute Error (MAE)
            & $\displaystyle
            \frac{1}{m} \sum_{i=1}^{m} \left| y_i - \hat{y}_i \right|
            $ \\

            Mean Squared Error (MSE)
            & $\displaystyle
            \frac{1}{m} \sum_{i=1}^{m} (y_i - \hat{y}_i)^2
            $ \\
            \bottomrule
            \end{tabular}
        \end{table}

    \subsection{Comparative Methods}
        To rigorously evaluate the CoNN framework, we compared it with two representative paradigms for inverse design of HPC. First is the standalone autoencoder variants, and second is the Bayesian inference method based on GP regression. These baselines were chosen to distinguish the effects of the cooperative learning mechanism from those of probabilistic search-based inference, providing a balanced comparison between modern neural and classical Bayesian methodologies.

        \subsubsection{Standalone Autoencoder}
            \label{sec:comparative_methods_ae}
            The first baseline group corresponds to the CoNN architecture trained without cooperative feedback between the surrogate and imputation models. This includes three configurations which are the Denoising Autoencoder (DAE), Denoising Variational Autoencoder (DVAE), and Denoising Wasserstein Autoencoder (DWAE). These methods utilize the same network backbones and training hyperparameters as the full CoNN but omit (i) the mask matrix ($\mathbf{MM}$) at the autoencoder input and in the reconstruction loss, (ii) feature clipping, and (iii) the dual loss coupling with the surrogate model, as these components constitute key contributions of the proposed framework. At inference, the mask still composes the final design from fixed and reconstructed variables as depicted in \figref{fig:myfigure4}.

            By comparing these non-cooperative variants to the fully cooperative CoNN, we can directly assess how much performance improvement arises from the feedback interaction between the surrogate (performance evaluator) and the imputation model (design generator). All models were trained and tested on identical data partitions with the same optimization settings to ensure fair comparison.

        \subsubsection{Bayesian Inference with Gaussian Process Surrogate}
            \label{sec:comparative_methods_bi}
            The second baseline is based on the Bayesian inference framework originally proposed by \cite{ke_bayesian_2021} for the full inverse design of HPC. Although their approach was not specifically developed for partial inverse design, it provides a probabilistic formulation that can be naturally extended to cases where only part of the input composition is unknown. The method combines a GP surrogate model with MCMC sampling via the MH algorithm, enabling probabilistic inference of mix designs consistent with a target property.

            In the context of partial inverse design, the design variable vector $\mathbf{X}$ can be partitioned into two subsets, $\mathbf{X} = [\mathbf{X}_{f}, \mathbf{X}_{u}]$, where $\mathbf{X}_{f}$ represents the fixed (specified) variables and $\mathbf{X}_{u}$ denotes the unknown variables to be inferred. The inverse problem can then be expressed in Bayesian form as:

            \begin{equation}
            p(\mathbf{X}_{u} \mid y^{*}, \mathbf{X}_{f}) \propto p(y^{*} \mid \mathbf{X}_{u}, \mathbf{X}_{f}) \cdot p(\mathbf{X}_{u})
            \label{eq:eq6}
            \end{equation}

            here, $p(y^{*} \mid \mathbf{X}_{u}, \mathbf{X}_{f})$ is the likelihood modeled by the GP surrogate, and $p(\mathbf{X}_{u})$ represents the prior distribution of the unknown variables. The posterior distribution $p(\mathbf{X}_{u} \mid y^{*}, \mathbf{X}_{f})$ quantifies the set of feasible mix designs that satisfy the target property under the given fixed-variable constraints. The likelihood term from the GP surrogate is typically modeled as a Gaussian distribution:

            \begin{equation}
            p(y^{*} \mid \mathbf{X}_{u}, \mathbf{X}_{f})
            = \mathcal{N}\!\left( f(\mathbf{X}_{u}, \mathbf{X}_{f}), \, \sigma^{2}(\mathbf{X}_{u}, \mathbf{X}_{f}) \right)
            \label{eq:eq7}
            \end{equation}

            where $f(\mathbf{X}_{u}, \mathbf{X}_{f})$ is the GP-predicted mean compressive strength, and $\sigma^{2}(\mathbf{X}_{u}, \mathbf{X}_{f})$ is the predictive variance.

            The MCMC implementation follows the standard Metropolis-Hastings procedure. The prior $p(\mathbf{X}_u)$ on each unknown variable is taken as a uniform distribution over the per-feature empirical range of the standardized training data, which prevents the chain from proposing values outside the support observed during surrogate training. Proposals are generated by a symmetric Gaussian random walk applied only to the unknown components, with the fixed components $\mathbf{X}_f$ reassigned to their user-specified values at every step so that the operator-imposed constraints are held exactly constant throughout sampling. Proposal components that fall outside the per-feature prior bounds are clipped to the bounds rather than rejected, preserving the symmetry of the random walk. Based on preliminary tuning of the Bayesian-GP baseline, we used the best-performing configuration for the proposal standard deviation and burn-in length. Specifically, the first 500 iterations were discarded as burn-in, and the proposal standard deviation was kept fixed for the remaining iterations.

            The MCMC sampling procedure explores this posterior distribution to identify candidate compositions. For fair comparison with the proposed CoNN, the Bayesian-GP method was reimplemented under identical data partitions and constraint settings. To match the single-design-per-query output of the proposed CoNN, the final mix design for each Bayesian-GP test case was computed as the posterior mean of the post-burn-in samples. During inference, sampling was performed with search sizes of 1, 100, 1,000, 10,000, and 100,000 post-burn-in MH searches for each test case. While this Bayesian approach can produce statistically valid and uncertainty-aware designs, it requires extensive iterative sampling to reach competitive performance. In contrast, the proposed CoNN generates comparable or superior results in a single forward pass, achieving higher computational efficiency and practical usability for partial inverse design.

%% file: manuscript-sec-results.tex
\section{Results and Discussion}
\label{sec:results}
    \subsection{Comparison with Standalone Models}
        \label{sec:comp_standalone}
        In our first quantitative comparison, we analyze the generated design candidates of the CoNN framework against standalone models, including DAE, DVAE, and DWAE, as explained in \secref{sec:experiment}.

\begin{figure}[htbp]
            \centering
            \includegraphics[width=1.0\textwidth]{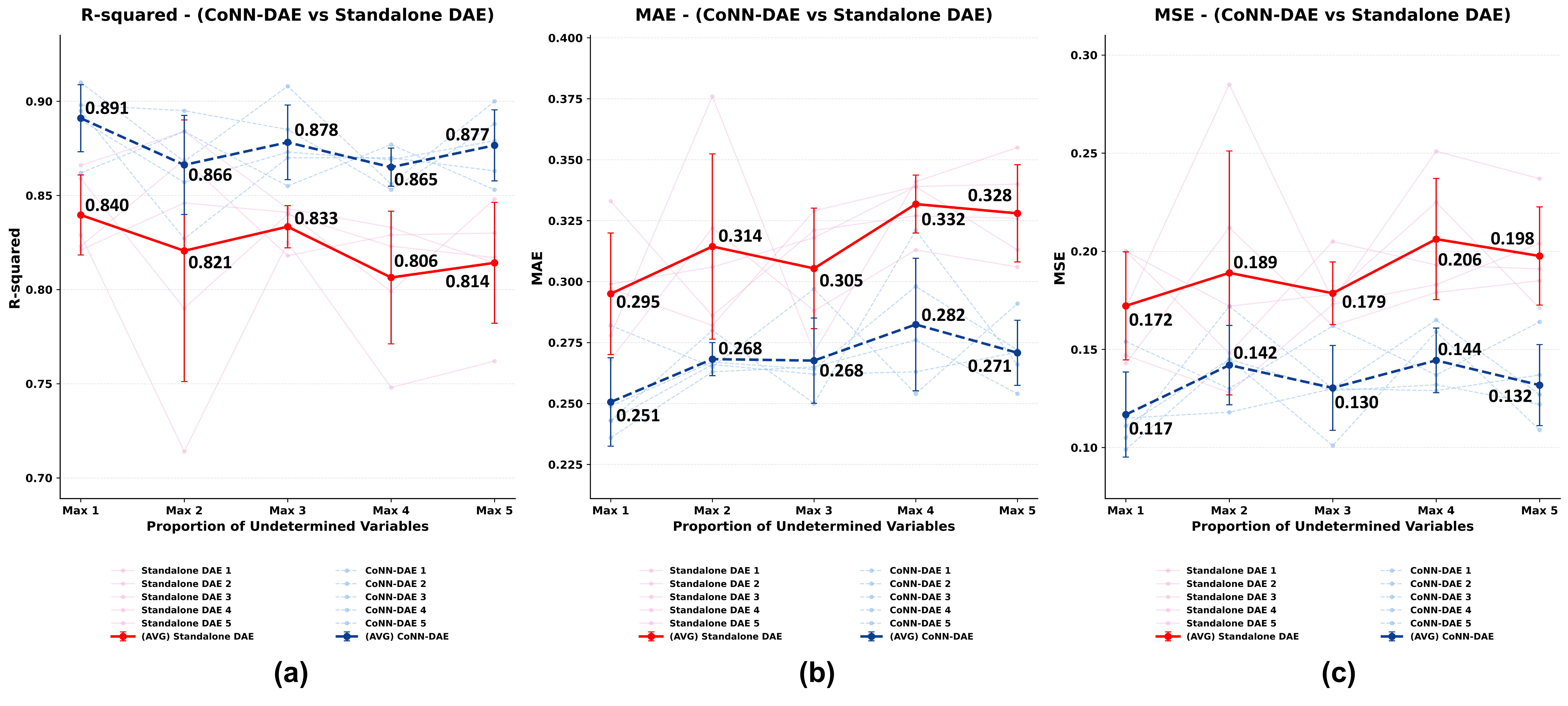}
            \caption{Performance comparison of standalone DAE and CoNN-DAE for (a) $R^2$, (b) MAE, and (c) MSE across varying proportions of undetermined variables. Average results over five runs ($n = 5$) are shown with $\pm 1$ standard deviation.}
            \label{fig:myfigure5}
        \end{figure}

\begin{figure}[htbp]
            \centering
            \includegraphics[width=0.9\textwidth]{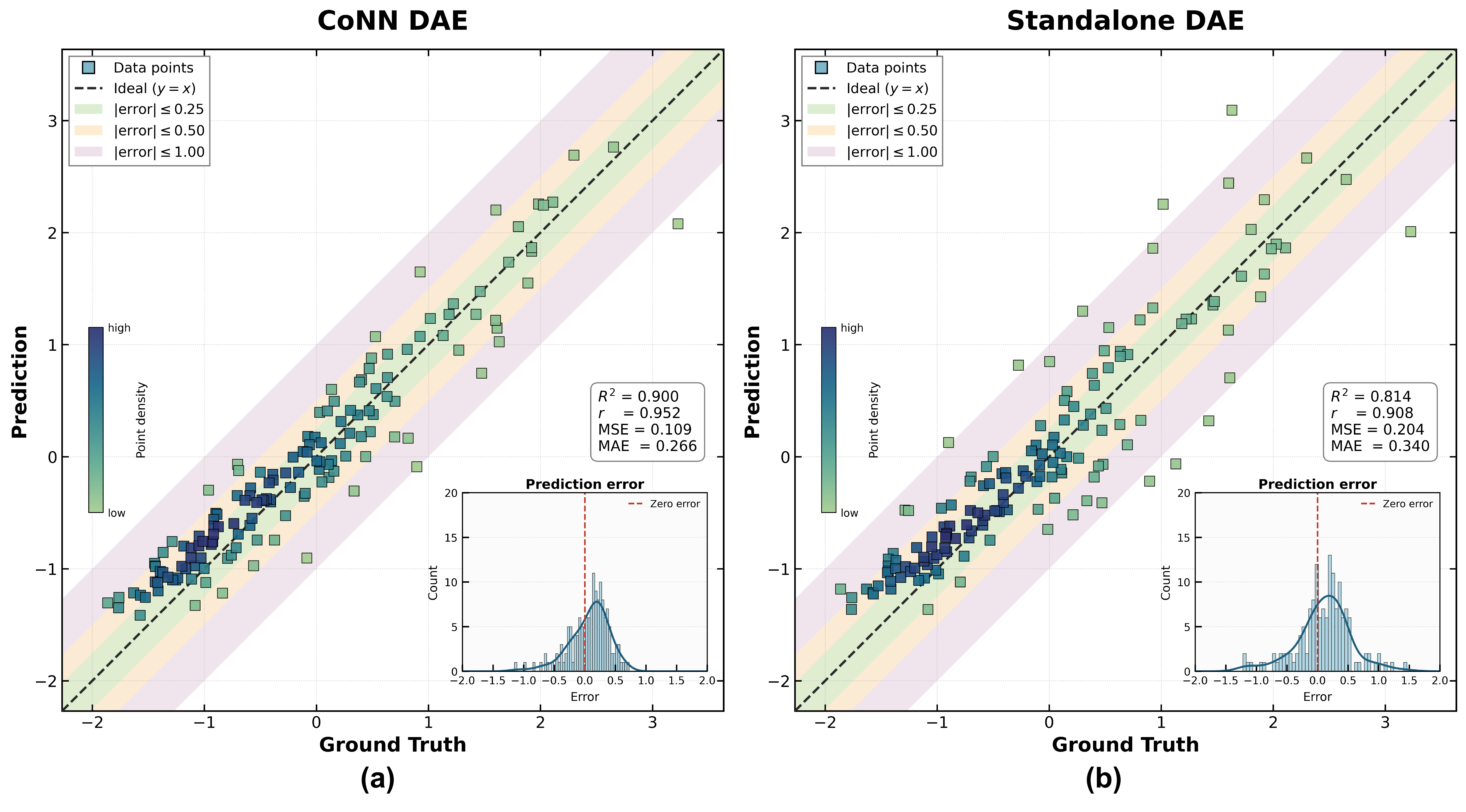}
            \caption{Prediction against ground truth for (a) CoNN-DAE and (b) Standalone DAE at the most challenging setting (maximum five undetermined variables), shown for the data split where the CoNN-DAE achieved its best test performance, with the standalone DAE evaluated on the same split for a paired comparison.}
            \label{fig:myfigure6}
        \end{figure}

        First, we evaluate the performance of the CoNN-DAE against the standalone DAE, as shown in \figref{fig:myfigure5}. The results demonstrate that the CoNN framework consistently outperforms its standalone counterpart. Aggregated across five independent splits, the CoNN-DAE delivers more stable results with lower variance, whereas the standalone DAE exhibits instability. This creates a significant performance gap, particularly as the number of undetermined variables increase, the CoNN-DAE remains stable while the standalone DAE degrades. Specifically, the CoNN-DAE achieves higher and more stable $R^2$ values in the range of 0.865-0.891, while the standalone DAE stays clearly lower and fluctuates between 0.806 and 0.840 with an overall decline from 0.840 to 0.814. The error metrics also highlight this advantage. The CoNN-DAE maintains lower MAE between 0.251-0.282, compared to 0.295-0.332 for the standalone DAE, and lower MSE between 0.117-0.144, whereas the standalone DAE fluctuates at a much higher error level and rises overall from 0.172 to 0.198. Under the most challenging scenario with maximum five undetermined variables, the CoNN-DAE reduces MAE by 17.4\% (from 0.328 to 0.271) and MSE by 33.3\% (from 0.198 to 0.132), demonstrating its robustness in partial inverse design tasks. \tabref{tab:tab3} and \figref{fig:myfigure6} show this comparison in detail for the single data split where the CoNN-DAE achieved its best test performance, while the averaged results across all five splits are presented in \figref{fig:myfigure5} and in the significance analysis of \tabref{tab:sig_standalone}. CoNN-DAE approaches the performance ceiling set by its underlying surrogate model. Moreover, the surrogate's accuracy is on par with the forward prediction results reported in the original work of \citet{yeh_modeling_1998} and the recent study by \citet{ke_bayesian_2021}, confirming that the cooperative framework delivers competitive performance within the present data-driven setting.

\begin{figure}[htbp]
            \centering
            \includegraphics[width=1.0\textwidth]{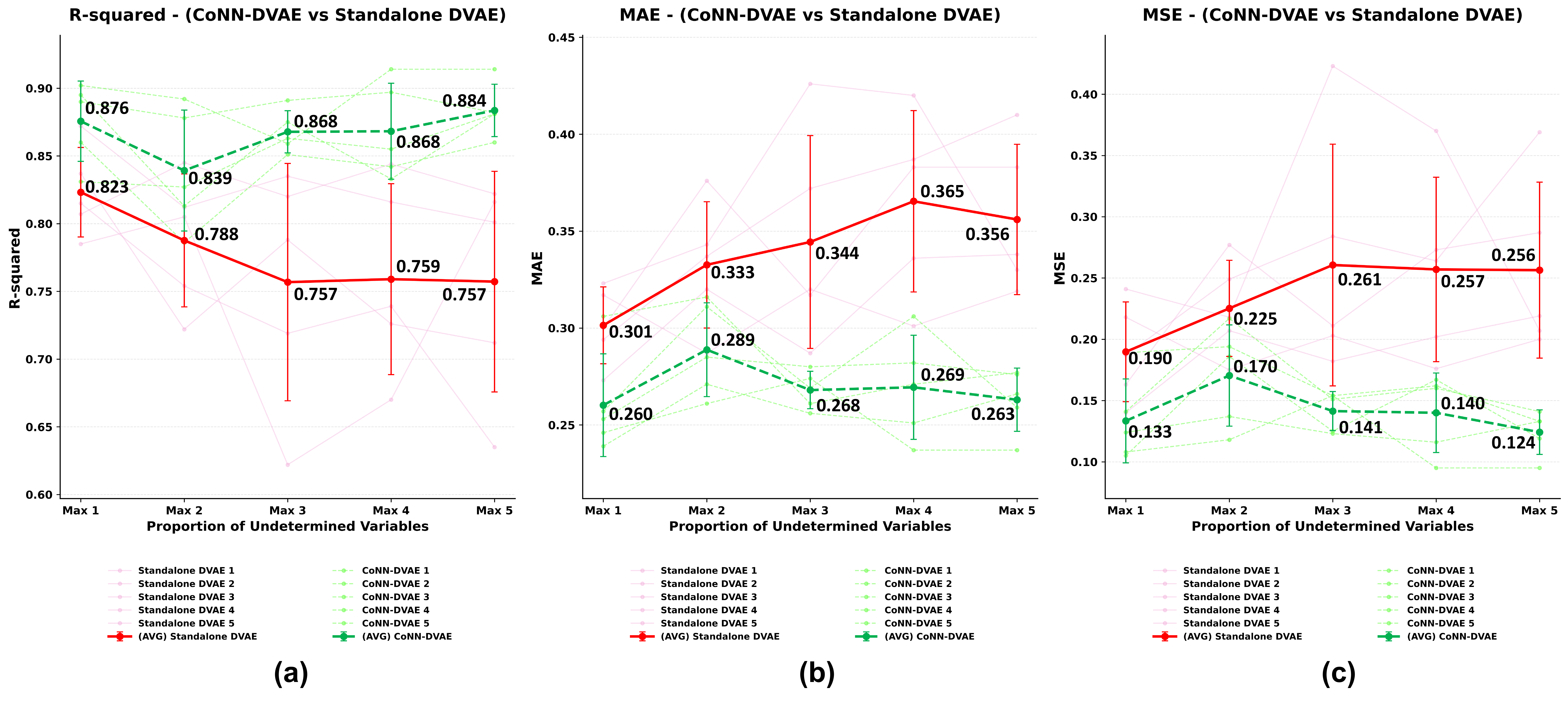}
            \caption{Performance comparison of standalone DVAE and CoNN-DVAE for (a) $R^2$, (b) MAE, and (c) MSE across varying proportions of undetermined variables. Average results over five runs ($n = 5$) are shown with $\pm 1$ standard deviation.}
            \label{fig:myfigure7}
        \end{figure}

        Next, we compare the CoNN-DVAE with the standalone DVAE, as shown in \figref{fig:myfigure7}. The results clearly indicate that the CoNN framework substantially improves both accuracy and stability. The CoNN-DVAE achieves consistently higher $R^2$ values, in the range 0.839-0.884, while the standalone DVAE declines from 0.823 to 0.757 as the number of undetermined variables increases. Similarly, the CoNN-DVAE maintains lower MAE values in the range 0.260-0.289, compared to 0.301-0.365 for the standalone DVAE. The advantage is even more evident in MSE, where the CoNN-DVAE stays between 0.124-0.170, while the standalone DVAE increases from 0.190 to 0.256. Under the most challenging condition with maximum five undetermined variables, the CoNN-DVAE reduces MAE by 26.1\% (from 0.356 to 0.263) and MSE by 51.6\% (from 0.256 to 0.124).

        \begin{table}[htbp]
            \centering
            \caption{Comparison of the surrogate model, CoNN-DAE, and standalone DAE on the most difficult scenario (maximum of 5 undetermined variables), reported for the same best-performing data split as \figref{fig:myfigure6}.}
            \label{tab:tab3}
            \renewcommand{\arraystretch}{1.25}
            \begin{tabular}{lccc}
            \toprule
            \textbf{Evaluation Setting} & \textbf{MAE} & \textbf{MSE} & \textbf{R\textsuperscript{2}} \\
            \midrule
            Surrogate Performance (mean) & 0.252 & 0.099 & 0.909 \\
            CoNN-DAE                     & 0.266 & 0.109 & 0.900 \\
            Standalone DAE               & 0.340 & 0.204 & 0.814 \\
            \bottomrule
            \end{tabular}
        \end{table}

\begin{figure}[htbp]
            \centering
            \includegraphics[width=1.0\textwidth]{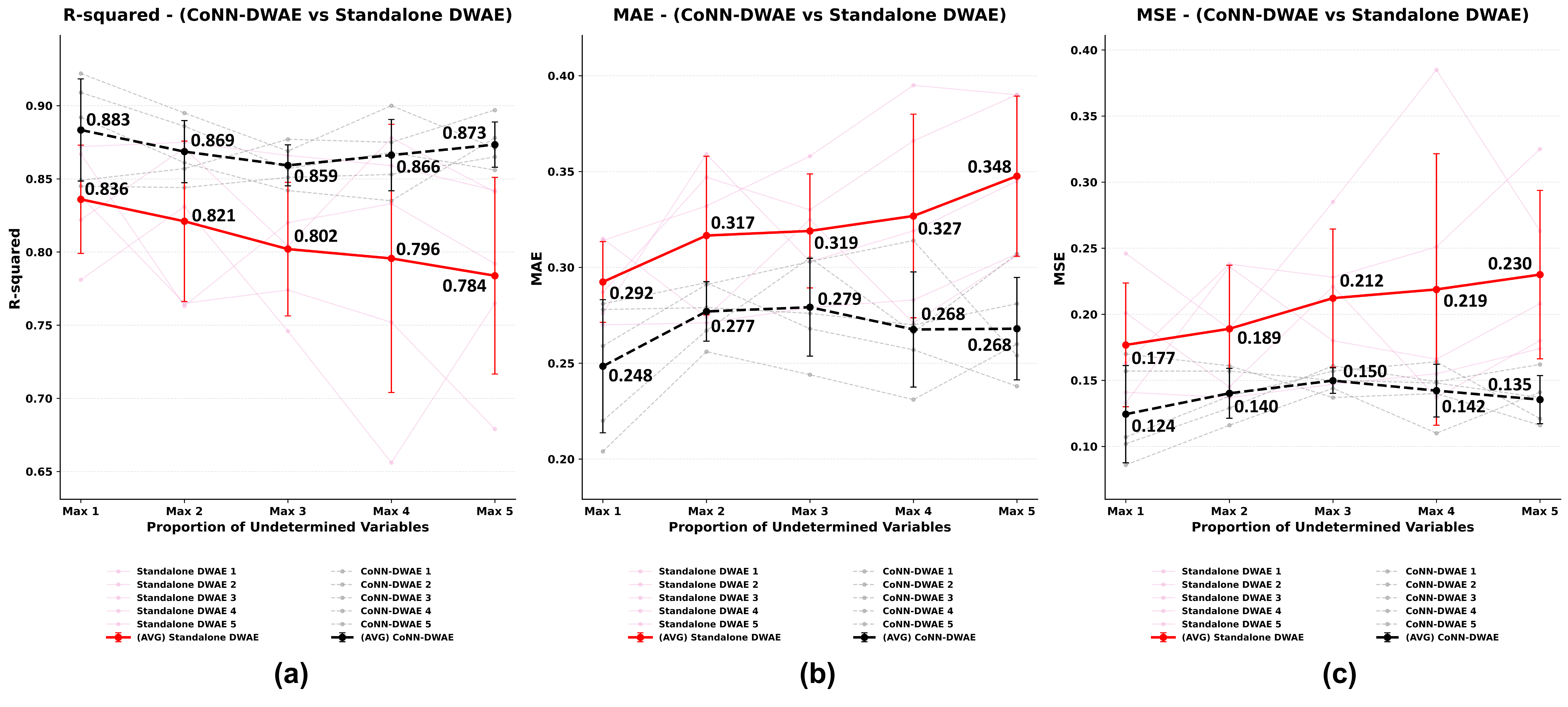}
            \caption{Performance comparison of standalone DWAE and CoNN-DWAE for (a) $R^2$, (b) MAE, and (c) MSE across varying proportions of undetermined variables. Average results over five runs ($n = 5$) are shown with $\pm 1$ standard deviation.}
            \label{fig:myfigure8}
        \end{figure}

        Finally, we compare the CoNN-DWAE with the standalone DWAE, as shown in \figref{fig:myfigure8}. Similar to the previous cases, the CoNN framework consistently demonstrates superior performance and robustness. The CoNN-DWAE achieves stable $R^2$ values in the range 0.859-0.883, while the standalone DWAE declines from 0.836 to 0.784 as the number of undetermined variables increases. In terms of error metrics, the CoNN-DWAE maintains lower MAE in the range 0.248-0.279, compared to 0.292-0.348 for the standalone DWAE. Likewise, the CoNN-DWAE achieves lower MSE in the range 0.124-0.150, whereas the standalone DWAE shows a rise in prediction error from 0.177 to 0.230. Under the most challenging scenario with maximum five undetermined variables, the CoNN-DWAE reduces MAE by 22.9\% (from 0.348 to 0.268) and MSE by 41.1\% (from 0.230 to 0.135).

        Across all experiments, a consistent pattern is observed. Models within the CoNN framework are more robust and exhibit lower variance compared to their standalone counterparts, which are substantially less stable and prone to large fluctuations. This difference is also evident across the five independent data splits, where CoNN models consistently yield tightly grouped results, while the standalone models show large variations from split to split, indicating instability and sensitivity to data partitioning. As task difficulty increases with a greater number of undetermined variables, the CoNN models remain stable, while standalone models suffer from declining performance and increasing error rates. This stability and reduced variance are particularly important in concrete research, where reliable and reproducible predictions are critical for guiding concrete mix design under practical constraints. Furthermore, the different CoNN variants serve complementary roles. The non-generative CoNN-DAE is well-suited for one-to-one mapping tasks such as reproducing high-performance concrete mix designs, while the generative CoNN-DVAE and CoNN-DWAE are more appropriate for one-to-many mapping tasks, enabling broader exploration of the feasible design space.

        To confirm that these observed performance gaps are statistically significant rather than artifacts of experimental randomness, we applied the paired significance tests described in \secref{sec:eval_objective} to the per-condition metric values. Across the $n=25$ matched pairs per metric, every cooperative variant outperforms its standalone counterpart on MSE, MAE, and $R^2$, with paired $t$-test $p$-values below $2{\times}10^{-5}$ and Wilcoxon signed-rank $p$-values below $3{\times}10^{-5}$, as summarized in \tabref{tab:sig_standalone}. The cooperative variants win in at least 23 of the 25 matched configurations for every metric and baseline, confirming that the observed advantage is consistent across random splits and masking configurations and is unlikely to be driven by seed-level variability alone.

        \begin{table}[htbp]
            \centering
            \caption{Paired significance test results for Cooperative NN (CoNN) variants versus their Standalone counterparts, computed over $n = 25$ matched pairs per metric (five random splits $\times$ five masking configurations). $\Delta$ denotes the mean improvement of CoNN over the baseline, where a positive value indicates that CoNN outperforms the baseline.}
            \label{tab:sig_standalone}
            \renewcommand{\arraystretch}{1.25}
            \setlength{\tabcolsep}{5pt}
            \begin{tabular}{llcccccc}
            \toprule
            \textbf{Comparison} & \textbf{Metric} & \textbf{CoNN} & \textbf{Standalone} & $\boldsymbol{\Delta}$ & \textbf{Wins} & $\boldsymbol{p_t}$ & $\boldsymbol{p_w}$ \\
            \midrule
            \multirow{3}{*}{CoNN-DAE vs Standalone DAE}
                          & MSE                  & 0.1331 & 0.1887 & $+0.0556$ & 24/25 & $1.9{\times}10^{-8}$  & $3.0{\times}10^{-7}$ \\
                          & MAE                  & 0.2679 & 0.3149 & $+0.0470$ & 25/25 & $2.3{\times}10^{-8}$  & $1.2{\times}10^{-5}$ \\
                          & $R^2$ & 0.8754 & 0.8228 & $+0.0526$ & 24/25 & $3.5{\times}10^{-8}$  & $3.0{\times}10^{-7}$ \\
                        \midrule
            \multirow{3}{*}{CoNN-DVAE vs Standalone DVAE}
                          & MSE                  & 0.1419 & 0.2378 & $+0.0959$ & 25/25 & $9.4{\times}10^{-8}$  & $6.0{\times}10^{-8}$ \\
                          & MAE                  & 0.2699 & 0.3400 & $+0.0701$ & 25/25 & $1.4{\times}10^{-8}$  & $1.2{\times}10^{-5}$ \\
                          & $R^2$ & 0.8669 & 0.7768 & $+0.0901$ & 25/25 & $9.8{\times}10^{-8}$  & $6.0{\times}10^{-8}$ \\
                        \midrule
            \multirow{3}{*}{CoNN-DWAE vs Standalone DWAE}
                          & MSE                  & 0.1384 & 0.2054 & $+0.0670$ & 24/25 & $1.9{\times}10^{-5}$  & $4.2{\times}10^{-7}$ \\
                          & MAE                  & 0.2680 & 0.3205 & $+0.0524$ & 23/25 & $3.7{\times}10^{-6}$  & $2.1{\times}10^{-5}$ \\
                          & $R^2$ & 0.8702 & 0.8077 & $+0.0625$ & 24/25 & $1.7{\times}10^{-5}$  & $2.0{\times}10^{-5}$ \\
            \bottomrule
            \end{tabular}
        \end{table}

    \subsection{Comparison with Bayesian Inference}
        In this section, we compare the CoNN framework, including CoNN-DAE, CoNN-DVAE, and CoNN-DWAE, against Bayesian inference with a GP surrogate. As Bayesian inference is a well-established baseline for inverse design, especially in concrete research, this evaluation highlights the advantages of the proposed approach. Both predictive accuracy and computational efficiency are assessed in the following subsections.

        \subsubsection{Prediction Accuracy}
            \label{sec:comp_bayesian}
\begin{figure}[htbp]
                \centering
                \includegraphics[width=1.0\textwidth]{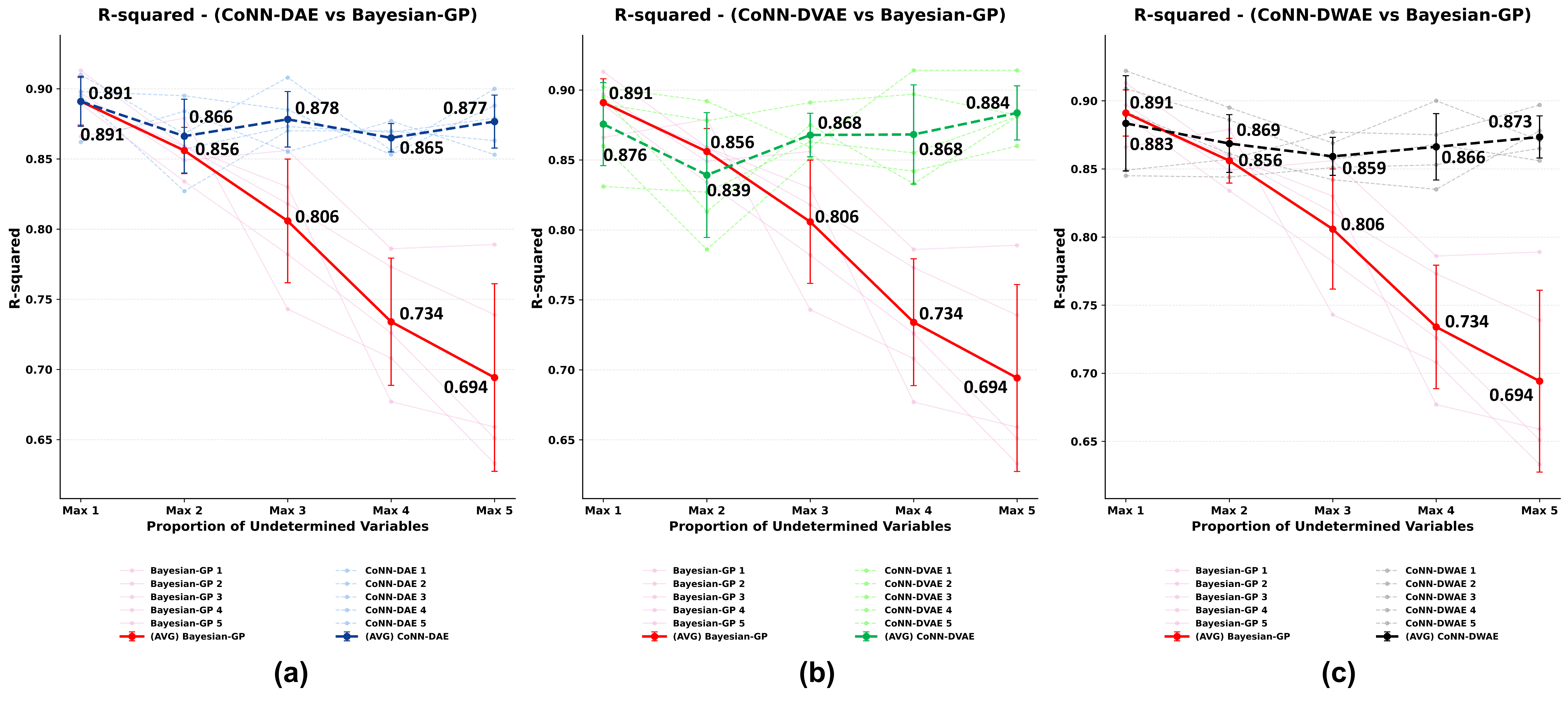}
                \caption{Comparative $R^2$ performance of Bayesian-GP (100{,}000 MH searches) with (a) CoNN-DAE, (b) CoNN-DVAE, and (c) CoNN-DWAE across varying proportions of undetermined variables. Average results over five runs ($n = 5$) are shown with $\pm 1$ standard deviation.}
                \label{fig:myfigure9}
            \end{figure}

            At the smallest proportion of undetermined variables (Max 1), all three CoNN variants perform at a level comparable to Bayesian inference, but they clearly outperform Bayesian inference as the proportion of undetermined variables increases, as shown in \figref{fig:myfigure9}. The CoNN models maintain higher and stable $R^2$ values around 0.839-0.891 across different levels of undetermined variables, showing only moderate fluctuations. In contrast, Bayesian inference starts at a similar $R^2$ value of 0.891 but deteriorates sharply as the number of undetermined variables increases, reaching as low as 0.694 at the most difficult setting. Under this condition, the CoNN framework demonstrates a relative improvement of approximately 26\% in $R^2$ compared to Bayesian inference, underscoring its superior predictive accuracy and robustness under high design uncertainty.

            \begin{figure}[htbp]
                \centering
                \includegraphics[width=1.0\textwidth]{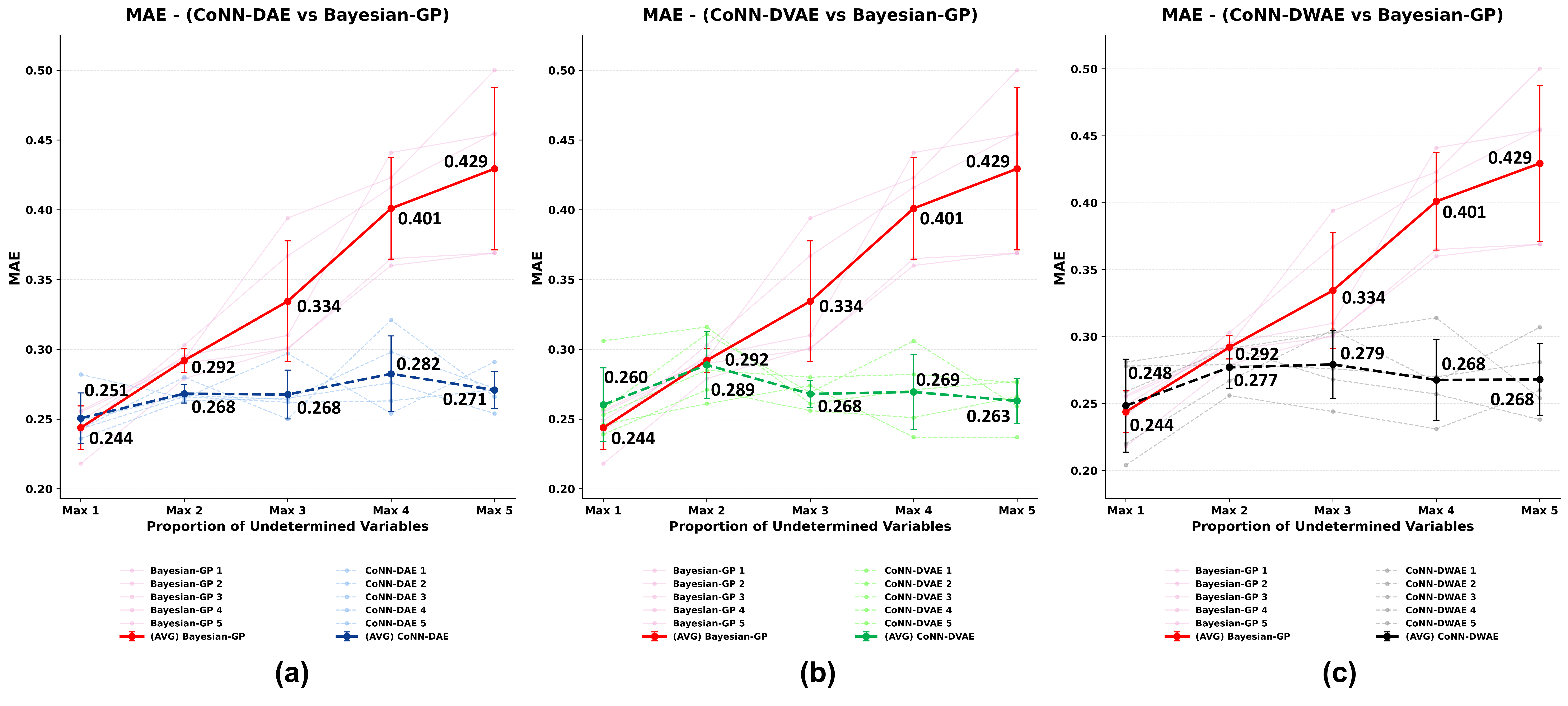}
                \caption{Comparative MAE performance of Bayesian-GP (100{,}000 MH searches) with (a) CoNN-DAE, (b) CoNN-DVAE, and (c) CoNN-DWAE across varying proportions of undetermined variables. Average results over five runs ($n = 5$) are shown with $\pm 1$ standard deviation.}
                \label{fig:myfigure10}
            \end{figure}
            
\begin{figure}[htbp]
                \centering
                \includegraphics[width=1.0\textwidth]{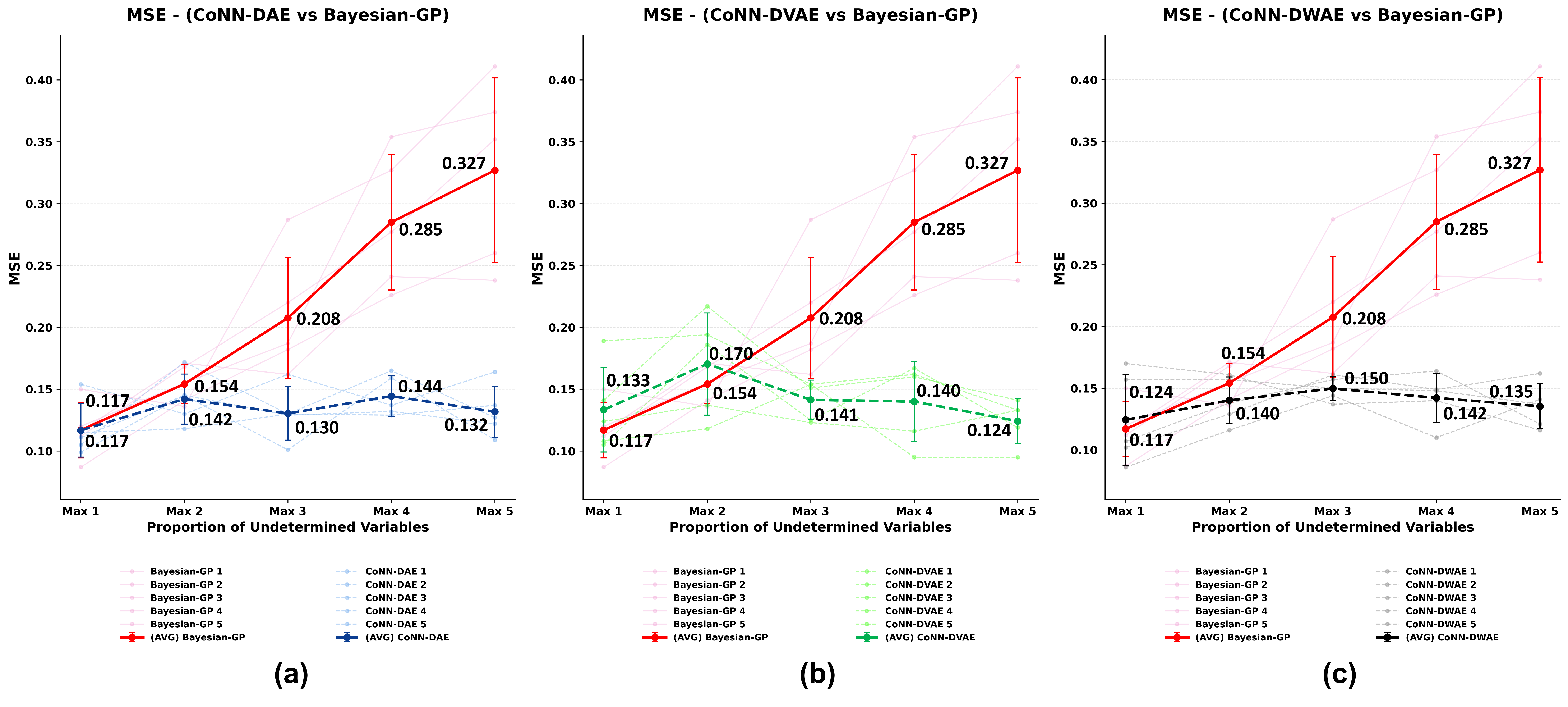}
                \caption{Comparative MSE performance of Bayesian-GP (100{,}000 MH searches) with (a) CoNN-DAE, (b) CoNN-DVAE, and (c) CoNN-DWAE across varying proportions of undetermined variables. Average results over five runs ($n = 5$) are shown with $\pm 1$ standard deviation.}
                \label{fig:myfigure11}
            \end{figure}

            The error-based evaluation further emphasizes the superiority of the CoNN framework over Bayesian inference, as shown in \figsref{fig:myfigure10} and ~\ref{fig:myfigure11}. For MAE, all CoNN models maintain relatively low errors with moderate fluctuations across different levels of undetermined variables, with values ranging from approximately 0.248 to 0.289. In contrast, Bayesian inference starts at a similar level of approximately 0.244 at the easiest setting but exhibits steadily increasing errors, rising to around 0.429 at the most difficult setting. A similar pattern is observed in the MSE results. The CoNN models keep their errors between approximately 0.117 and 0.170 across all difficulty levels, while Bayesian inference starts at a similar level of about 0.117 but climbs to approximately 0.327 as task difficulty grows. At the most challenging level, the CoNN framework achieves substantial absolute error reductions compared to Bayesian inference, reducing MAE by approximately 37.7\% and MSE by about 60.1\%. These results confirm that the CoNN models not only deliver significantly lower error rates but also maintain relatively consistent performance as the number of undetermined variables increases, whereas Bayesian inference consistently deteriorates with higher task difficulty.

            Taken together, the $R^2$ and error-based evaluations clearly demonstrate the superiority of the CoNN framework over Bayesian inference. Across all scenarios, the CoNN models remain stable with low variance, while Bayesian inference shows a consistent decline in predictive accuracy and a steady increase in error as task difficulty grows. Moreover, Bayesian inference produces highly inconsistent outcomes across the five independent splits, with large fluctuations between runs despite using the same experimental setup. This instability underscores its sensitivity to data partitioning and highlights the difficulty of obtaining reliable results with sampling-based approaches. Such stability is particularly important in concrete research, where reproducible and reliable predictions are essential for guiding practical mix designs under uncertainty. By combining accurate forward prediction with robust handling of incomplete design information, the CoNN framework establishes itself as a more effective and dependable approach than Bayesian inference for partial inverse design tasks.

            Applying the same paired significance procedure to this comparison confirms that the advantage of the cooperative framework over Bayesian inference is statistically significant. Across the $n = 25$ matched pairs per metric, every CoNN variant outperforms the Bayesian-GP baseline on MSE, MAE, and $R^2$, with paired $t$-test $p$-values below $2{\times}10^{-3}$ and Wilcoxon signed-rank $p$-values below $3{\times}10^{-3}$, and every CoNN variant winning in at least 18 of the 25 matched configurations per metric, as summarized in \tabref{tab:sig_bayesian}. Combined with the descriptive comparisons above, these results confirm that the accuracy gap between the cooperative framework and Bayesian inference is a statistically supported and consistent effect rather than a visual impression from the mean curves.

            \begin{table}[htbp]
                \centering
                \caption{Paired significance test results for Cooperative NN (CoNN) variants versus the Bayesian-GP baseline (100{,}000 MH searches), computed over $n = 25$ matched pairs per metric (five random splits $\times$ five masking configurations). $\Delta$ denotes the mean improvement of CoNN over the baseline, where a positive value indicates that CoNN outperforms the baseline.}
                \label{tab:sig_bayesian}
                \renewcommand{\arraystretch}{1.25}
                \setlength{\tabcolsep}{5pt}
                \begin{tabular}{llcccccc}
                \toprule
                \textbf{Comparison} & \textbf{Metric} & \textbf{CoNN} & \textbf{Bayesian-GP} & $\boldsymbol{\Delta}$ & \textbf{Wins} & $\boldsymbol{p_t}$ & $\boldsymbol{p_w}$ \\
                \midrule
                \multirow{3}{*}{CoNN-DAE vs Bayesian-GP}
                      & MSE   & 0.1331 & 0.2182 & $+0.0851$ & 22/25 & $5.4{\times}10^{-5}$  & $1.0{\times}10^{-4}$ \\
                      & MAE   & 0.2679 & 0.3401 & $+0.0722$ & 23/25 & $2.5{\times}10^{-5}$  & $7.2{\times}10^{-5}$ \\
                      & $R^2$ & 0.8754 & 0.7962 & $+0.0792$ & 22/25 & $5.4{\times}10^{-5}$  & $1.1{\times}10^{-4}$ \\
                                \midrule
                \multirow{3}{*}{CoNN-DVAE vs Bayesian-GP}
                      & MSE   & 0.1419 & 0.2182 & $+0.0763$ & 18/25 & $1.1{\times}10^{-3}$  & $2.6{\times}10^{-3}$ \\
                      & MAE   & 0.2699 & 0.3401 & $+0.0702$ & 20/25 & $2.8{\times}10^{-4}$  & $9.3{\times}10^{-4}$ \\
                      & $R^2$ & 0.8669 & 0.7962 & $+0.0707$ & 18/25 & $1.1{\times}10^{-3}$  & $2.6{\times}10^{-3}$ \\
                                \midrule
                \multirow{3}{*}{CoNN-DWAE vs Bayesian-GP}
                      & MSE   & 0.1384 & 0.2182 & $+0.0798$ & 21/25 & $3.1{\times}10^{-4}$  & $4.2{\times}10^{-4}$ \\
                      & MAE   & 0.2680 & 0.3401 & $+0.0721$ & 19/25 & $2.3{\times}10^{-4}$  & $2.9{\times}10^{-4}$ \\
                      & $R^2$ & 0.8702 & 0.7962 & $+0.0740$ & 21/25 & $3.0{\times}10^{-4}$  & $1.4{\times}10^{-4}$ \\
                \bottomrule
                \end{tabular}
            \end{table}

        \subsubsection{Computational Efficiency}
            \label{sec:comp_efficiency}
            In addition to predictive performance, computational efficiency was evaluated to assess the practicality of the inverse-design frameworks. We measure efficiency as the inference time. This is the wall-clock time needed to generate mix designs for the entire held-out test partition of 149 samples, with each method running on the hardware it is designed for. The CoNN framework runs on a GPU and produces all 149 designs at once with a single forward pass, because neural networks can process many designs in parallel. In contrast, the Bayesian-GP baseline is a sequential sampling procedure that runs on a CPU and must repeat its full Metropolis-Hastings search separately for every single design, following the evaluation protocol described in \secref{sec:comparative_methods_bi}. This ability to use modern parallel hardware is a genuine practical benefit of the neural network approach and one of the reasons for the large gap in inference time.

\begin{figure}[htbp]
                \centering
                \includegraphics[width=0.8\textwidth]{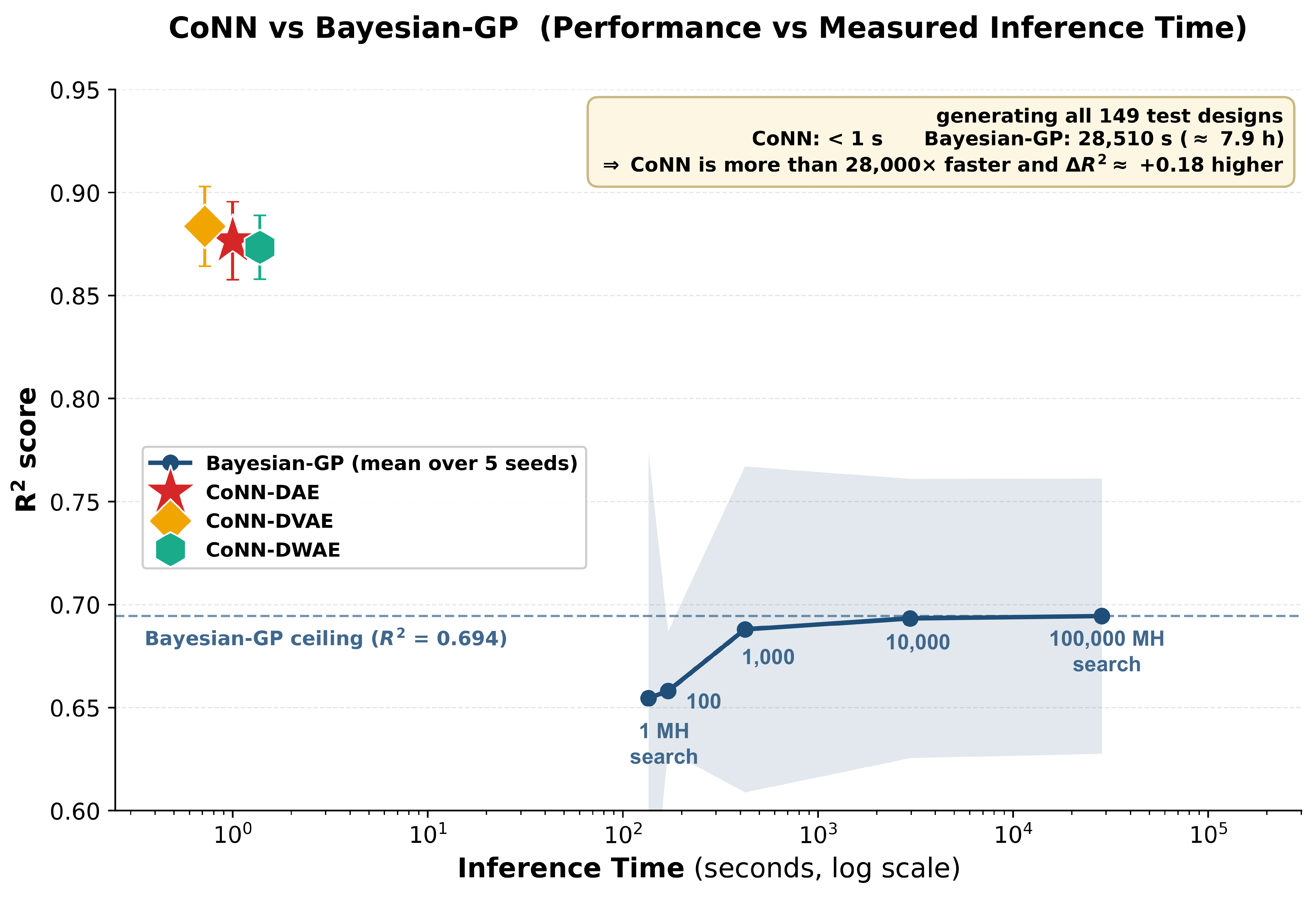}
                \caption{Comparison of $R^2$ score against measured inference time at the most challenging setting (maximum five undetermined variables), evaluated on the held-out test partition, for CoNN (single forward pass, three variants) and Bayesian-GP (1 to 100{,}000 MH searches for each design, mean $\pm$ 1 SD over five seeds).}
                \label{fig:myfigure12}
            \end{figure}

            \figref{fig:myfigure12} shows the $R^2$ score as a function of the measured inference time at the most challenging setting (maximum five undetermined variables), evaluated on the held-out test partition described in \secref{sec:datasplit}. The Bayesian-GP curve is plotted for five search sizes, with 1, 100, 1{,}000, 10{,}000, and 100{,}000 post-burn-in MH searches, and is reported as the mean over the five random seeds with a $\pm 1$ SD shaded band. The three CoNN variants (CoNN-DAE, CoNN-DVAE, and CoNN-DWAE) appear together at their measured time of less than one second, with vertical error bars over the same five seeds. A dashed reference line marks the Bayesian-GP performance ceiling at the largest search size evaluated.

            The contrast is substantial. The three CoNN variants generate all 149 test designs in less than one second and attain $R^2$ in the range 0.87 to 0.88, with a mean of 0.878. The Bayesian-GP baseline needs about 28{,}510 seconds (roughly 7.9 hours) at its largest search size. Its accuracy plateaus around $R^2 = 0.694$ and shows no further meaningful improvement beyond 10{,}000 MH searches. The CoNN framework is therefore more than 28{,}000 times faster and still exceeds the Bayesian-GP plateau by an absolute margin of $\Delta R^2 \approx +0.18$, which is a relative improvement of approximately 26\% in $R^2$. Analogous absolute gaps on MAE and MSE at the same setting are already established in the descriptive comparison of \secref{sec:comp_bayesian}, where $\Delta$MAE $\approx -0.16$ and $\Delta$MSE $\approx -0.20$.

            Such capability is especially critical for practical applications in concrete research and engineering practice, where real-time or large-scale mix-design exploration is often required under strict resource constraints. The single-pass inference of the CoNN framework keeps the cost of one design essentially constant, in contrast to the Bayesian-GP baseline whose long search must be repeated from scratch for every new constraint configuration. By combining accurate forward prediction with near-instant inference, the CoNN framework provides a scalable solution that makes partial inverse design feasible for deployment in real-world scenarios. We note that wall-clock times depend on the hardware and the implementation, so the reported times are indicative rather than absolute.

    \subsection{Perspective on Model Application}
        \label{sec:perspective}
        To demonstrate the applicability of the proposed framework in practical scenarios, \tabref{tab:tab6} presents an example of generating an HPC mix design while meeting performance requirements under controlled design variables. In this case, a sample from the test set is used where the undetermined variables are cement, water, coarse aggregate, and fine aggregate, while all other variables are predetermined. The objective is to achieve the target compressive strength of 48.4 MPa. Although this example focuses specifically on these four variables, the framework is inherently flexible and can be applied to different subsets of variables without retraining, enabling random or scenario-specific selections depending on design requirements, as outlined in the previous section.

        The results clearly highlight the superiority of the CoNN framework. The CoNN models achieved by far the lowest deviations between the predicted and target strengths, with differences of only $-$0.75 MPa, $-$2.91 MPa, and $-$1.08 MPa for CoNN-DAE, CoNN-DVAE, and CoNN-DWAE, respectively. By contrast, the standalone models and Bayesian-GP missed the target by a much larger margin, with deviations ranging from $-$11.23 MPa to $-$18.00 MPa. It is also worth noting that the cement content generated by CoNN-DWAE (540.00 kg/m$^3$) lies exactly at the upper bound of the training data. This shows the clipping step in action, as it actively kept the generated design inside the feasible range. These outcomes confirm that the CoNN framework not only generates feasible mix designs but also more reliably fulfills the target strength criteria provided by the user, which is essential for ensuring long-term durability in practical applications.

        From a methodological perspective, the proposed framework is positioned at the candidate-generation stage of a data-driven inverse-design pipeline for engineering materials. It is intended to support design-space exploration in the pre-qualification of HPC mixtures under user-defined constraint patterns. Therefore, its main role is to generate constraint-aware candidate mixes for subsequent bench-scale verification, rather than to produce finalized engineering specifications. The present study introduces and benchmarks the framework using a single public HPC dataset. Accordingly, applying the framework to a new HPC material system or a new dataset requires use-case-specific retraining with a representative dataset, followed by external validation using independent HPC mix-design records.

        \begin{table}[htbp]
            \centering
            \caption{Application and comparison of generated design results from all models. The estimated strength of the CoNN and standalone rows is computed by the ANN surrogate model, while the Bayesian-GP row is computed by its own GP surrogate.}
            \label{tab:tab6}
            \renewcommand{\arraystretch}{1.25}
            \setlength{\tabcolsep}{3.3pt}

            \begin{tabularx}{\textwidth}{l l *{8}{c} c}
            \toprule
            \multirow{2}{*}{\textbf{Category}}
            & \multirow{2}{*}{\textbf{Model}}
            & \multicolumn{8}{c}{\textbf{Design variables}}
            & \multirow{2}{*}{\shortstack[c]{\textbf{Estimation}\\\textbf{strength (diff)}}} \\
            \cmidrule(lr){3-10}
            & & \textbf{Cement} & \textbf{BFS} & \textbf{PFA} & \textbf{Water}
            & \textbf{SP} & \textbf{CA} & \textbf{FA} & \textbf{Age} &  \\
            \midrule

            \textbf{User input}
            & --
            & \dots & 0.00 & 0.00 & \dots & 0.00 & \dots & \dots & 14.00
            & \textbf{48.40} \\
            \midrule

            \multirow{7}{*}{\shortstack[c]{\textbf{Modelling}\\\textbf{results}}}
            & Bayesian-GP
            & \textbf{362.36} & 0.00 & 0.00 & \textbf{180.62} & 0.00 & \textbf{975.96} & \textbf{772.36} & 14.00
            & \textbf{32.82 (-15.58)} \\

            & Standalone DAE
            & \textbf{374.52} & 0.00 & 0.00 & \textbf{177.05} & 0.00 & \textbf{1061.96} & \textbf{705.73} & 14.00
            & \textbf{30.40 (-18.00)} \\

            & Standalone DVAE
            & \textbf{496.57} & 0.00 & 0.00 & \textbf{199.19} & 0.00 & \textbf{1026.08} & \textbf{662.07} & 14.00
            & \textbf{37.17 (-11.23)} \\

            & Standalone DWAE
            & \textbf{438.79} & 0.00 & 0.00 & \textbf{193.75} & 0.00 & \textbf{955.71} & \textbf{652.57} & 14.00
            & \textbf{36.77 (-11.64)} \\

            & Cooperative DAE
            & \textbf{505.14} & 0.00 & 0.00 & \textbf{171.75} & 0.00 & \textbf{909.34} & \textbf{744.18} & 14.00
            & \textbf{47.65 (-0.75)} \\

            & Cooperative DVAE
            & \textbf{476.46} & 0.00 & 0.00 & \textbf{160.29} & 0.00 & \textbf{1002.04} & \textbf{833.65} & 14.00
            & \textbf{45.49 (-2.91)} \\

            & Cooperative DWAE
            & \textbf{540.00} & 0.00 & 0.00 & \textbf{146.22} & 0.00 & \textbf{873.73} & \textbf{698.71} & 14.00
            & \textbf{47.32 (-1.08)} \\
            \bottomrule
            \end{tabularx}
        \end{table}

%% file: manuscript-sec-conclusion.tex
\section{Conclusion}
\label{sec:conclusion}
    This study applied a Cooperative Neural Networks (CoNN) framework to address the challenge of partial inverse design in the cement and concrete domain, marking its first application to HPC. Unlike conventional forward or inverse design approaches, partial inverse design allows only a subset of variables to remain undetermined while others are fixed by design constraints. The framework, composed of an imputation model and a surrogate model trained under cooperative learning, enables the generation of mix designs under flexible constraints with a single forward pass.

    The experimental results demonstrate several key contributions. First, the CoNN framework consistently outperformed standalone autoencoder variants (DAE, DVAE, and DWAE), achieving higher predictive accuracy, lower error rates, and substantially reduced variance as the number of undetermined variables increased. This robustness is particularly valuable in practical scenarios, where stable and reproducible results are critical despite rising task complexity. Second, when benchmarked against Bayesian inference with GP surrogates, the CoNN models not only achieved superior accuracy but also demonstrated remarkable computational efficiency, producing valid mix designs within a single forward pass that takes less than one second. This efficiency stands in sharp contrast to Bayesian approaches that require thousands of iterative searches and hours of computation time, further underscoring the practicality and scalability of the CoNN framework for inverse design in concrete research.

    Despite these promising results, several limitations remain. The dataset used in this study is relatively scarce, which may limit the generalizability of the findings compared to scenarios where larger and more diverse datasets are available. At the same time, because the CoNN framework couples the imputation model with a surrogate predictor, its highest achievable performance is inherently bounded by the predictive fidelity of that surrogate, so improving surrogate accuracy through larger and more diverse training data is itself an important direction for future work. Beyond expanding the training data, future work could further address this challenge by exploring more advanced strategies such as neural networks that capture physics information such as PINN, which enable domain knowledge to be injected into the learning process through common-sense or policy-driven shape constraints (e.g., absolute-volume balance, water--binder ratio bounds, etc.), thereby better capturing the physical and mechanistic relationships governing HPC mix design.

    In summary, this research pioneers the application of CoNN for partial inverse design in HPC, demonstrating strong predictive accuracy and computational efficiency. With further refinements in data availability and methodological integration, the framework has the potential to evolve into a powerful tool for intelligent, constraint-aware material design.

%% file: manuscript-sec-acknowledgements.tex
\section*{Acknowledgements}
    This research was supported by the MSIT (Ministry of Science and ICT), Korea, under the ICAN (ICT Challenge and Advanced Network of HRD) support program supervised by the IITP (Institute for Information \& Communications Technology Planning \& Evaluation) (IITP-2025-RS-2023-00259806).

%% file: manuscript-sec-appendix.tex
\appendix

\section{Ablation of the min--max clipping step}
\label{sec:appendix_clipping}

    This appendix quantifies the effect of the empirical min--max feature clipping step in \eqnref{eq:eq3} on three concrete-specific feasibility checks, evaluated on the held-out test partition for the three imputation variants (DAE, DVAE, DWAE) after inverse-scaling to original units and rounding to four decimals, as summarized in \tabref{tab:clipping_ablation}. Binder denotes the total cementitious content, namely the sum of cement, blast-furnace slag (BFS), and fly ash (PFA). The water-to-binder ratio is defined as $w/b = \mathrm{Water} / (\mathrm{Cement} + \mathrm{BFS} + \mathrm{PFA})$ and checked against the empirical training envelope of $[0.2351,\, 0.9000]$. The supplementary cementitious materials (SCM) replacement ratio is defined as $\mathrm{SCM} = (\mathrm{BFS} + \mathrm{PFA}) / (\mathrm{Cement} + \mathrm{BFS} + \mathrm{PFA})$ and checked against $[0.0000,\, 0.7360]$. A degeneracy guard flags the pathological case of a non-positive binder total that would make the two ratios undefined.

    \begin{table}[htbp]
        \centering
        \caption{Violation rates and observed ranges of concrete-specific feasibility checks on generated mixes, before and after the per-feature min--max clipping step. Percentages and ranges are computed on the held-out 1CV test partition ($N = 149$) under the random-masking protocol with up to five masked variables per sample.}
        \label{tab:clipping_ablation}
        \begin{tabular}{@{}l l c c c c@{}}
            \toprule
            \multirow{2}{*}{\textbf{Variant}} & \multirow{2}{*}{\textbf{Check}} & \multicolumn{2}{c}{\textbf{Before clipping}} & \multicolumn{2}{c}{\textbf{After clipping}} \\
            \cmidrule(lr){3-4} \cmidrule(lr){5-6}
            & & \textbf{Viol. (\%)} & \textbf{Observed range} & \textbf{Viol. (\%)} & \textbf{Observed range} \\
            \midrule
            \multirow{3}{*}{CoNN-DAE}  & $w/b$ ratio      & 6.711 & $[0.2145,\, 1.4418]$ & 2.685 & $[0.2145,\, 1.1706]$ \\
                                       & SCM replacement  & 10.067 & $[-0.5412,\, 0.6880]$ & 0.000 & $[0.0000,\, 0.6880]$ \\
                                       & binder $\le 0$   & 0.000 & N/A                   & 0.000 & N/A \\
            \midrule
            \multirow{3}{*}{CoNN-DVAE} & $w/b$ ratio      & 7.383 & $[0.2495,\, 1.9403]$  & 1.342 & $[0.2495,\, 1.1319]$ \\
                                       & SCM replacement  & 8.054 & $[-1.6796,\, 0.7006]$ & 0.000 & $[0.0000,\, 0.6982]$ \\
                                       & binder $\le 0$   & 0.000 & N/A                   & 0.000 & N/A \\
            \midrule
            \multirow{3}{*}{CoNN-DWAE} & $w/b$ ratio      & 9.396 & $[0.2392,\, 1.6942]$  & 1.342 & $[0.2392,\, 1.2813]$ \\
                                       & SCM replacement  & 10.067 & $[-1.6961,\, 0.6928]$ & 0.000 & $[0.0000,\, 0.6928]$ \\
                                       & binder $\le 0$   & 0.000 & N/A                   & 0.000 & N/A \\
            \bottomrule
        \end{tabular}
    \end{table}

    The per-feature clipping step removes all SCM-replacement violations across all variants, and no non-positive binder cases occur. This is because keeping the individual material values within their training ranges is sufficient to satisfy these two checks when the binder remains positive. A small residual $w/b$ violation nevertheless persists after clipping, reflecting the structural limitation of axis-aligned clipping, namely that a compound ratio whose numerator and denominator are clipped independently is not itself guaranteed to remain within its empirical envelope. Hard-constraint enforcement of such compound ratios is identified in \secref{sec:conclusion} as a target for physics-informed extensions of the cooperative framework.